\pdfoutput=1

\documentclass[11pt]{article}

\usepackage[final]{ACL2023}

\usepackage{times}
\usepackage{latexsym}
\usepackage{boldline}
\usepackage{caption}
\usepackage{algorithm}
\usepackage{algorithmic}
\usepackage{multirow}
\usepackage{amssymb}
\usepackage{pifont}
\usepackage{tikz}

\usepackage{graphicx} 
\usepackage[T1]{fontenc}

\usepackage[utf8]{inputenc}

\usepackage{microtype}

\usepackage{inconsolata}

\title{EDeR: A Dataset for Exploring Dependency Relations Between Events}

\author{
 Ruiqi Li$^{\spadesuit}$, \ \
 Patrik Haslum$^{\spadesuit}$, \ \
 Leyang Cui$^{\heartsuit}$ \\ 
 $^\spadesuit$ Australian National University\\
 $^\heartsuit$ Tencent AI lab \\
 \textit{\{ruiqi.li,\ patrik.haslum\}@anu.edu.au} \quad\textit{leyangcui@tencent.com}\\
}

\begin{document}
\maketitle
\begin{abstract}
Relation extraction is a central task in natural language processing (NLP) and information retrieval (IR) research. We argue that an important type of relation not explored in NLP or IR research to date is that of an event being an argument -- required or optional -- of another event. We introduce the human-annotated {\bf E}vent {\bf De}pendency {\bf R}elation dataset (EDeR) which provides this dependency relation.
The annotation is done on a sample of documents from the OntoNotes dataset, which has the added benefit that it integrates with existing, orthogonal, annotations of this dataset. We investigate baseline approaches for predicting the event dependency relation, the best of which achieves an accuracy of 82.61\% for binary argument/non-argument classification. We show that recognizing this relation leads to more accurate event extraction (semantic role labelling) and can improve downstream tasks that depend on this, such as co-reference resolution. Furthermore, we demonstrate that predicting the three-way classification into the required argument, optional argument or non-argument is a more challenging task.
\end{abstract}

\section{Introduction}
 Events play a crucial role in enabling AI agents to understand and perceive the world, as they provide information on what happened and the entities involved.
 An event is composed of a predicate (i.e., verb) and arguments, where the predicate indicates the event's action and the arguments represent the subject, object and so on of the predicate~\cite{levin1999objecthood,hovav2010lexical}. Representing and extracting dependency relations among events that can identify both semantic (e.g., tell what the object of an event predicate is) and syntactic (e.g., tell what the relative clause(s) of an event predicate is) relationships is an interesting task for AI researchers. Such event dependency relations also provide useful information that benefits some downstream tasks in Information Retrieval (IR) and Natural Language Processing (NLP).
 
 \begin{figure}[!t]
\centering
\includegraphics[width=0.48\textwidth]{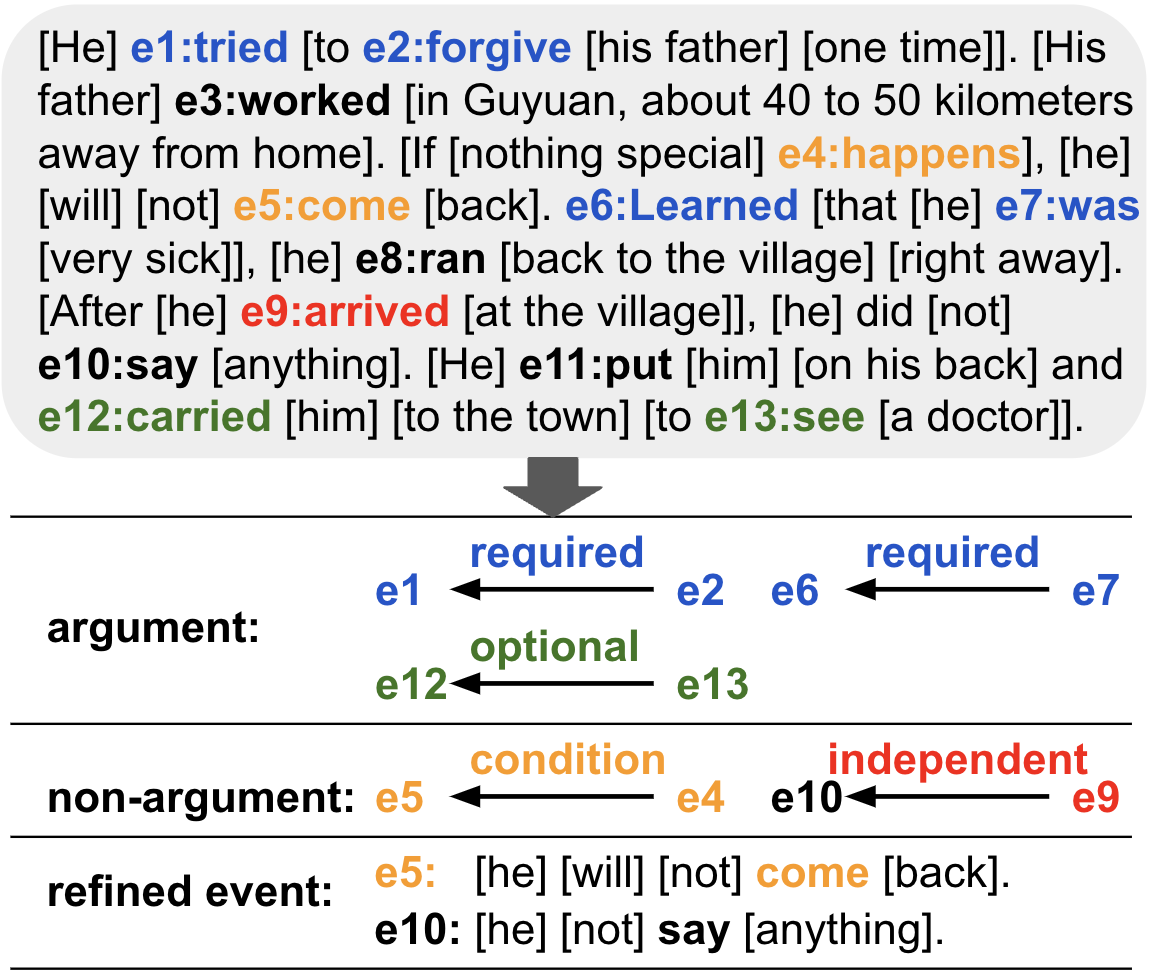}
\caption{Examples of the event dependency relations. Above: Source text. (Event predicates are marked in boldface and argument spans are marked with brackets.) Below: Event dependency relations and refined events. argument: event pairs with dependency relation and whether it is required or optional; non-argument: event pairs with no dependency relations and whether the right-hand event is a condition of or independent from the left-hand event; refined events: the left-hand events from the non-argument event pairs with refined argument spans.}
\label{fig-intro}
\end{figure}

Previous work on relation extraction in IR and NLP domains mainly focused on investigating the relationships between entities ~\cite{miwa-sasaki-2014-modeling,huguet-cabot-navigli-2021-rebel-relation,chen-etal-2020-joint,ma-etal-2022-joint} rather than events. In the relatively limited research work that studies the relations between events, the types of relations considered are causal~\cite{mirza-tonelli:COLING:2016, mariko:arXiv:2020, mariko-etal-2022-financial, yang:arXiv:2022}, temporal~\cite{bethard:SemEval:2013,laokulrat:SemEval:2013,chambers:2013,chambers:ACL:2014,mirza-tonelli:COLING:2016}, and hierarchical \cite{hovy2013events,glavas-etal-2014-hieve,glavas-snajder-2014-constructing,araki2014detecting}.

Complementing these, we investigate the relation of an event being an \textbf{argument} of another event, as opposed to being \textbf{independent}.
For example, in the sentence ``He tried to forgive his father one time.'' from Figure~\ref{fig-intro}, $e2$ with the predicate ``forgive'' is the action that he ``tried'' ($e1$) to take.
However, there is no temporal or causal or hierarchical relation between
$e2$ and $e1$; in fact, the sentence implies that $e2$ (``forgive'') did not actually occur.
We further distinguish between \textbf{required argument} and \textbf{optional argument} events. A required argument must be present for an event to be complete and meaningful. For example, in Figure~\ref{fig-intro}, $e2$ (``forgive'') must serve as an argument of $e1$ (``tried''); if removed, $e1$ becomes just ``He tried to'', which is incomplete and not meaningful (to ``try to'', one has to try to do something). An optional argument only adds to or clarifies the event that it is an argument of, for example indicating the purpose.
In Figure \ref{fig-intro}, $e13$ with the predicate ``see'' is such an optional argument of $e12$ (``carried''). Although optional, it has the property of an argument event that the sentence does not state that it actually occurs.
In contrast, a non-argument event is one that is conjunctive or conditional. However, such non-argument events are often mislabelled as arguments in event extraction (e.g., by semantic role labelling (SRL) systems). For example, event $e9$ (``arrived''), although contained in a temporal modifier of event $e10$ (``did not say anything''), is independent.
Argument-event relations are frequent in natural language texts, especially in narrative texts such as news articles, conversations, and similar.
We contend that identifying this dependency relation between events leads to a more accurate extraction of events, and of other relations between them.

Motivated by these, we present a human-annotated {\bf E}vent {\bf De}pendency {\bf R}elation dataset (EDeR) that (1) extracts event dependency information based on a sample of documents from OntoNotes~\cite{pradhan2013towards} and integrates with orthogonal annotations of this dataset, and (2) provides refined semantic role-labelled event representations based on this information.

We adopt a multi-level quality inspection-based approach to annotate event pairs from 275 articles under 7 different genres from the OntoNotes dataset. In the end, we build an event dependency relation dataset (EDeR) that contains 11,852 high-quality annotated samples, according to the proposed event relation taxonomy.
It can serve as the basis for an effective automatic classification process. We employ various text classification models, from creating heuristic rules to fine-tuning the Transformer~\cite{vaswani2017attention}-based pre-trained models. The binary argument/non-argument classification results establish solid baseline performances in predicting the event dependency relationships.

Furthermore, we apply both the EDeR human-annotated and the outperformed baseline model predicted event dependency relations to guide a SOTA SRL model~\cite{zhang-etal-2022-crfsrl} for event representation generation. The experimental results reveal that our event dependency relation can improve the SRL model's performance in extracting the updated event representations that our EDeR dataset provides. We further apply the event representations to a downstream task: co-reference resolution (CR). Comparing the performance of the SOTA CR model~\cite{jiang-cohn-2021-coref-hgat} using the original event representations to the performance of the model using our updated event representations proves the plausibility and validity of the refined event representations from our EDeR dataset. 

In addition, we also investigate the baseline performances for the three-way (i.e., required argument, optional argument or non-argument) classification, the experimental results indicate that it is a more challenging task for these language models. We release EDeR and source code for baselines at \url{https://github.com/RichieLee93/EDeR}.

\section{Background and Related Work}

\subsection{Event and Event Representation}
An event mention consists of a verb or phrasal verb as the predicate and a set of labelled arguments ~\cite{levin1999objecthood,hovav2010lexical}. Events can be extracted from texts by performing semantic role labelling (SRL) to label the roles (i.e., predicates and arguments) of words or word spans in a given sentence.
The OntoNotes dataset follows the PropBank annotation schema ~\cite{bonial:journal:2012} for semantic role labels, in which arguments are text spans, and which divides argument labels into numbered arguments (ARG0--ARG5), for arguments required for the valency of an action (e.g., agent and patient), and modifiers of the verb, such as purpose (PRP), locative (LOC), and so on. Examples are shown in the top part of Figure~\ref{fig-conv_ontonote_anno}.

There are also some public datasets (e.g., CoNLL-2005 and CoNLL-2012) for some shared SRL tasks (e.g., CoNLL 2005~\cite{carreras-marquez-2005-conll}, CoNLL 2009~\cite{hajic-etal-2009-conll}, and CoNLL 2012~\cite{pradhan-etal-2012-conll} tasks.).  CoNLL-2012 has been merged into the OntoNotes V5.0 dataset, so we call this dataset as OntnoNotes in the rest of the paper. 

An event can itself be an argument of another event. In the annotated SRL datasets, we find such argument events contained within the span of the event they are an argument of, but the converse does not hold: events that are contained in the span of another event are not always argument events, but may share part of their arguments (subject or object) with the containing event.
For example, as shown in Figure~\ref{fig-conv_ontonote_anno}, the event ``The man [who] works here'' is contained in the event ``The man who works here tells me to get the hell out'' but is not an argument of ``tells''. Such cases reveal that the annotation of events in these SRL datasets does not consider the argument event relation between events. Making this relation explicit helps us select a more focused argument span, such as ``The man'' for ARG0 (agent) of the event with the predicate ``tells'' in the example above.

\subsection{Relations Between Events}
Several event-event relations have been proposed for decades. TimeBank ~\cite{pustejovsky:book:2003} TB-Dense ~\cite{cassidy2014annotation} and MATRES ~\cite{Ning:ACL:2018} Dataset focus on the temporal relationship which reveals if one event happens BEFORE, AFTER, etc. another. Logical relationships like causality are also explored in datasets like CausalTimeBank ~\cite{mirza-tonelli:COLING:2016}, CiRA ~\cite{fischbach:etal:2021} and CNC ~\cite{tan:CNC:2022}. Such relationships regard two events as syntax and semantic agnostic - the occurrence and actuality of each event are isolated and not affected by others. 

Besides, several datasets propose hierarchical relationships (i.e., super-event and sub-event relations), where a sub-event should satisfy 2 conditions: Temporally, the sub-event occurs during the super-event; Spatially, the sub-event should be contained by the super-event ~\cite{hovy2013events,glavas-etal-2014-hieve,glavas-snajder-2014-constructing}. Araki et al.~\cite{araki2014detecting} introduces a dataset detecting event co-reference relation - whether two event mentions refer to the same event. Such hierarchical relations cannot deal with cases when an event (especially the verb of the event) requires a clausal complement, i.e., an argument of the event verb is itself an event, as the ``tried'' and ``forgive'' examples shown in Figure \ref{fig-intro}. 

\subsection{Event Relation Extraction}
Relation extraction is a popular branch of information retrieval research. Most recent works, as a downstream task of Named Entity Recognition (NER), concentrate on extracting relations between entities rather than events~\cite{miwa-sasaki-2014-modeling,huguet-cabot-navigli-2021-rebel-relation,chen-etal-2020-joint,ma-etal-2022-joint}.

As described, events are represented in word span style and such event representations can be regarded as text clips. Recently, there is remarkable progress in text relation classification tasks (e.g., textual entailment detection, sentiment analysis.) using Transformer~\cite{vaswani2017attention}-based language models like BERT, RoBERTa, GPT models~\cite{balazs-etal-2017-refining,ethayarajh-2019-contextual,laskar-etal-2020-contextualized,lee-etal-2022-toward}. Such language models are pretrained on a tremendous corpus, people further finetune it for their specific tasks using their selected datasets (so-called ``transfer learning''~\cite{pan-transfer}). This pretrain-finetune process usually achieves more advanced performance on relation extraction tasks, especially when the data size is small, compared with solely trained on a classical machine learning models or neural networks~\cite{balazs-etal-2017-refining,papanikolaou-etal-2019-deep,Seganti2021MultilingualEA}.
Learning from these, we can also adapt these state-of-the-art language models to classify the dependency relations between event pairs. Furthermore, based on different styles of event representations (i.e., semantic role-based~\cite{carreras-marquez-2005-conll,hajic-etal-2009-conll,pradhan-etal-2012-conll,pradhan2013towards} or predicate-based~\cite{pustejovsky:book:2003,cassidy2014annotation,Ning:ACL:2018,mirza-tonelli:COLING:2016}) and other features such as the syntactic dependency relation between predicates and so on, we can combine various inputs with the language models to explore the baseline performance on our EDeR dataset.

\section{Dataset}

As described in the last section, there is, as far as we are aware, no dataset of natural language text that identifies events and labels their argument dependency relation. We therefore construct such a dataset (named EDeR).
We use a subset of documents from OntoNotes, due to its available human-annotated and predicate-argument formatted events, and extract candidate event pairs, where the predicate of one event is contained in the span of the other, from these documents. Each event pair (i.e., $Event1$ and $Event2$) is labelled by human annotators to indicate whether $Event2$ is a required argument of, an optional argument of, or a condition of or independent from $Event1$.
We next detail the way we collect and pre-process the candidate event pairs and the human annotation process and construction of the dataset.

\subsection{Data Collection}

OntoNotes contains semantic role-formatted event representations, as the OntoNotes example in Figure~\ref{fig-conv_ontonote_anno} shows. We randomly sampled 275 documents from seven genres: broadcast news (bn), magazine (mz), newswire (nw), pivot corpus (pt), telephone conversation (tc), broadcast conversation (bc), and web data (wb). Data statistics of the 275 raw documents are shown in Table~\ref{tab-doc-info}. The number of sampled documents and the separation of them into training, development and test sets under each genre follows their initial distributions in the OntoNotes dataset.

\begin{table*}[!htbp]
\begin{center}
\begin{tabular}{lcccccccc} 
\textbf{} & \textbf{bn} & \textbf{mz} & \textbf{nw} & \textbf{pt} & \textbf{tc} & \textbf{bc}  & \textbf{wb} & \textbf{overall} \\
\hline
\# documents        & 104 & 7 & 107 & 31 & 7 & 3 & 16 & 275\\

\# documents-train       & 90 & 5 & 89 & 24 & 5& 1  & 13 & 227\\
\# documents-dev       & 7 & 1 & 9 & 4 & 1 & 1 & 2 & 25\\
\# documents-test       & 7 & 1 & 9 & 3 & 1 & 1 & 1 & 23\\
\hline
avg \# sentences per doc   & 6.2 & 48.9 & 14.4 & 40.7 & 71.7 & 203.8  & 59.3 & 24.9 \\
avg \# words per doc    & 139.7 & 1599.7 & 418.7 & 657.0 & 1089.4 & 4045.8 & 1509.3 & 543.4 \\
avg \# events per doc   &23.6  &202.1  &58.3  & 148.7 &362.1  &1058.6   & 268.5 & 93.5 \\

\hline

\end{tabular}
\caption{Statistics of the documents (under each genre and all) that we sampled from the OntoNotes dataset for annotation: number of documents and the number of documents split into different sets (train, development and test), and average number of sentences, words and events per document.}
\label{tab-doc-info}
\end{center} 
\end{table*}

\subsubsection{Candidate Event Pair Extraction}

\begin{figure}[!t]
\centering
\includegraphics[width=0.48\textwidth]{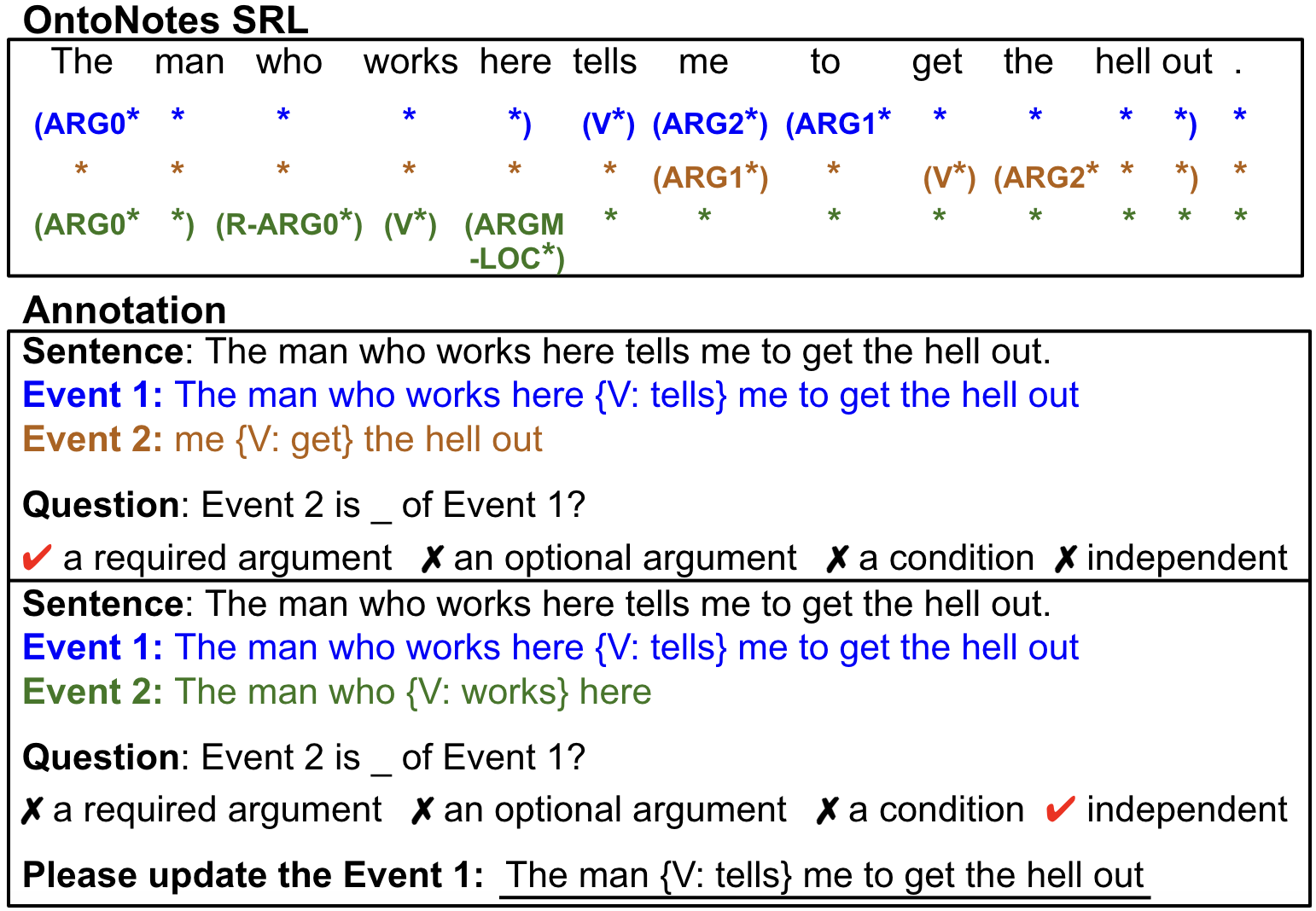}
\caption{Above: A sample sentence with semantic role labels from the OntoNotes dataset. Below: Corresponding event pairs presented for human annotation with event dependency relations.}
\label{fig-conv_ontonote_anno}
\end{figure}

Because arguments are spans of text, part or all of an extracted event may lie within the argument of another event. If the verb (i.e., predicate) of $e_j$ is within an argument of $e_i$, we say $e_j$ is \emph{contained} in $e_i$. This can be nested. Contained events are candidates for being argument events, but are not
necessarily so. For example, in the top part of Figure~\ref{fig-conv_ontonote_anno}, the event in green (\textbf{e3} = $\{$ARG0: The man $\}$ $\{$R-ARG0: who $\}$ $\{$V: works $\}$ $\{$ARGM-LOC: here $\}$) is contained in the span of the event in blue (\textbf{e1} = $\{$ARGO: The man who works here $\}$ $\{$V: tells $\}$ $\{$ARG2: me $\}$ $\{$ARG1: to get the hell out $\}$), but \textbf{e3} is not an argument of \textbf{e1}; however, the event in orange (\textbf{e2} = $\{$ARG1: me $\}$ $\{$V: get $\}$ $\{$ARG2: the hell out $\}$) is an argument of \textbf{e1}.

We select event pairs ($e_i$, $e_j$) where $e_j$'s verb is contained within the span of an argument of $e_i$ as candidate event pairs for annotation.

\subsubsection{Preprocessing}

We apply three filters to the selected candidate event pairs.
First, we filter out candidate event pairs ($e_i$, $e_j$) in which $e_j$ does not have any arguments. This typically occurs when $e_j$ is a modal verb indicating the tense of the verb in $e_i$.
Second, events in the OntoNotes dataset sometimes mistake adjectives for verbs (e.g., ``reloaded'' in ``the Matrix reloaded''), so we use the words' POS tags to filter these out. The POS tags of the event predicates are obtained using the Stanford CoreNLP toolkit \cite{manning:ACL:2014}.
Third, we remove transitively contained event pairs, i.e., pairs ($e_i$, $e_j$) such that there exists another event $e_k$ such that the verb of $e_k$ is contained in $e_i$ and the verb of $e_j$ is contained in $e_k$. In such cases, ($e_i$, $e_k$) and ($e_k$, $e_j$) may be candidate pairs, but ($e_i$, $e_j$) is not.
For example, in ``[She] $\{$V: try $\}$ [to $\{$V: stop $\}$ [[them] from $\{$V: ruining $\}$ [...]]]''), ``try'' and ``stop'' and ``stop'' and ``ruin'' are candidate pairs, but ``try'' and ``ruin'' are not.
 
After preprocessing, we obtain 11,852 candidate event pairs for the human annotation from the 275 documents.

\begin{figure}[!t]
\centering
\includegraphics[width=0.48\textwidth]{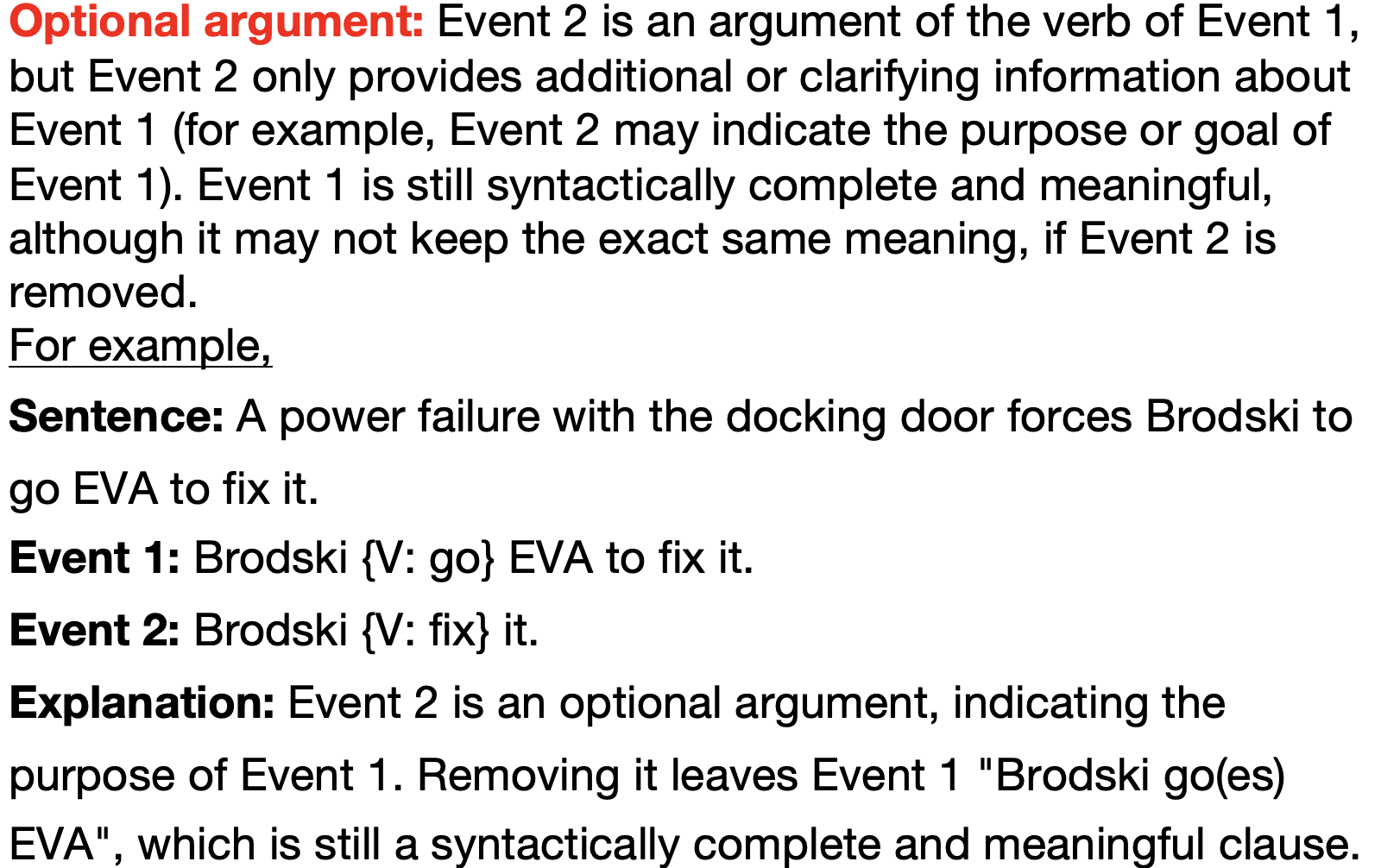}
\caption{A screenshot of part of the instructions for annotators. For each of the four labels, we provide the definition, examples of the relation, and a detailed explanation of why the example is labelled as it is. The screenshot shows this for the ``optional argument'' label.}
\label{fig-anntation-guideline}
\end{figure}

\subsection{Annotation}

\subsubsection{Annotation Instruction}

We present each candidate pair with the whole span of both events and highlight the predicate of each event with ``\{V: \}'', as shown in the lower part of Figure~\ref{fig-conv_ontonote_anno}. Annotators can choose one of four options to answer the relation between the two events.
For the annotation task, we separate the non-argument case into two: \textit{condition} and \textit{independent} (``not an argument or a condition''). This made the definitions of the labels easier to understand, since a condition is also, like argument events, a hypothetical, rather than actual, event in the text. Identifying conditional statements in text is itself an interesting and active topic of research ~\cite{fischbach:etal:2021,tan:CNC:2022}, motivated in particular by their relation to causality. However, the events labelled conditions in our dataset may only be a subset of all conditionals in the source text, because they are restricted to occurring in contained event pairs.
For the analysis of the dataset and experiments in the remainder of this paper, we treat the two non-argument labels as one.

We prepared annotation instructions including the task description, the definitions of the four labels, and 2--3 examples with a corresponding explanation of why the decision is made for each label, as shown Figure~\ref{fig-anntation-guideline}.

\subsubsection{Annotation Procedure}

\begin{figure}[!t]
\centering
\includegraphics[width=0.48\textwidth]{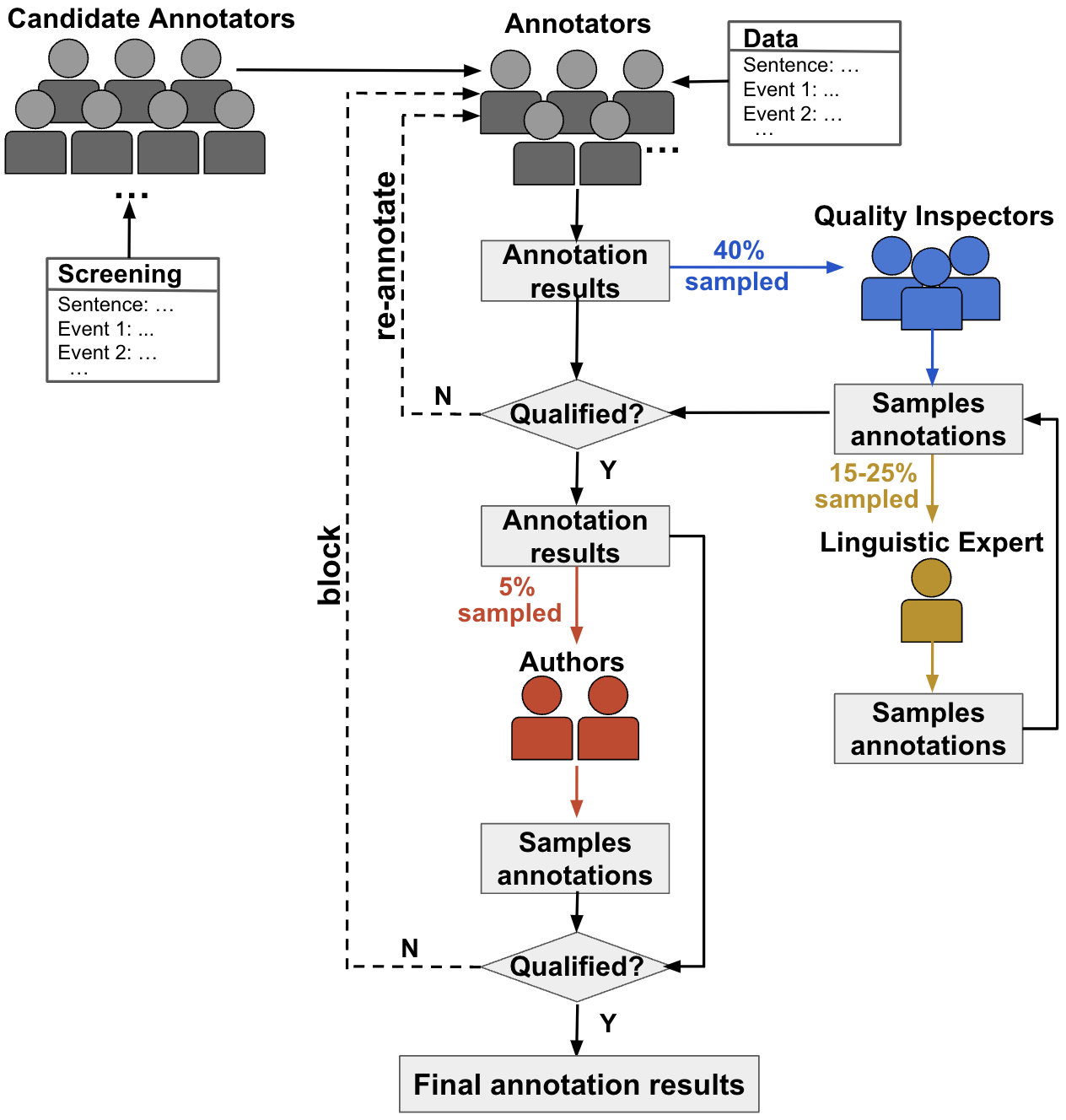}
\caption{Illustration of our multi-level qualification-based annotation procedure.}
\label{fig-annotate-procedure}
\end{figure}

We adopt a multi-level qualification-based annotation procedure in which annotators' work is sampled and inspected/corrected in several stages, and feedback from the inspections is passed back to the annotators. 
A schematic of the procedure is shown in Figure \ref{fig-annotate-procedure}.
Compared with the commonly-used crowd-sourcing and voting strategy, this procedure makes annotators learn to improve throughout the task. Although in the end not all event pairs have been annotated by multiple participants, the strict qualification tests used along the way ensure annotators give their best answers.

Over 80 \textit{candidate annotators} were recruited from colleges in China through a data annotation agency. They are English linguistic-major graduate students and fluent in English (passed the TEM-8 test\footnote{Test for English Majors-band 8, the highest level test in China for English major students that measures the overall English proficiency.}). After reading the instructions, candidates took a screening test, requiring them to annotate 50 examples, also annotated by authors, and cross-checked with the \textit{linguistic expert}. The linguistic expert who has rich NLP-related annotation experience was hired via the same data annotation agency. 16 candidates who answered at least 85\% of these questions correctly in the test were selected as \textit{annotators}. Among the 16 annotators, the three with the highest accuracy were selected as \textit{quality inspectors} (QIs). After further training, in which the linguistic expert explained the guideline and their mis-annotated cases from the screening test, the annotators started the annotation. 

The event pairs for annotation were evenly split into 3 subsets and sequentially released to the annotators in 3 stages, each one week apart. In each stage, after the annotators submit their annotations, the QIs randomly select and annotate 40\% of the cases. The expert sampled 15-25\% of cases inspected by the QIs, and checked the QIs' annotation results. If any incorrect annotations are found, the expert will send them back to the QIs with the expert's annotations and corresponding explanations to make the QIs understand the problems and avoid the same errors in the next stage. This kind of expert supervision controls the quality of the QIs and then avoids untrustable labels from QIs that mislead the annotators.
After the linguistic expert's review of the QIs' annotations, if the agreement between an annotator's answers and the corresponding QI's is lower than 95\%, the annotator was asked to re-annotate the inconsistent cases. The loop stops by when the agreement ratio reaches 95\%.

Two of the authors also monitor the quality of the annotation. In each stage, we sampled 5\% of the once-only annotated results from each annotator and labelled these cases to check their accuracy. If the accuracy achieved by an annotator is below 85\%, the work of that annotator was discarded, and the annotator removed from subsequent stages. Only one annotator failed this test and was removed after the first stage. Not counting the author inspection cases, 4771 (40.2\% of all) cases received at least four times annotation.

 \subsubsection{Event Representation Revision}

As described in the second annotation example in Figure~\ref{fig-conv_ontonote_anno}, one event (predicate) can be contained in another event but actually not an argument of it.
For the event pairs that annotators label as independent, we further ask them to revise containing event (``Event 1'') by removing the span of the contained event (``Event 2'') from it. However, the contained event often shares subject or object with the containing event, so annotators will keep any shared parts in Event 1 and remove the only remaining Event 2 spans.
An example is shown in Figure~\ref{fig-conv_ontonote_anno}: In the second case, Event 2, ``The man who \{V: works\} here'', is not an argument or a condition of Event 1, but shares the subject ``The man'' with Event 1. The refined Event 1 is ``The man \{V: tells\} me to get the hell out''.
Combined with the argument labels of the original event representations, the refined event representation becomes $\{$ARG0: The man $\}$ $\{$V: tells $\}$ $\{$ARG2: me $\}$ $\{$ARG1: to get the hell out $\}$. This manual event representation revision proceeds simultaneously with the annotation of the event dependency relation and is inspected by the QIs and the linguistic expert as well.

For events that are labelled as conditions, the contained event normally does not share subject or object with the containing event. We observed that we can often revise the containing event automatically, and therefore did not ask annotators to do so in this case.
If the conditional event $e_j$ and its containing event $e_i$ appear in the sentence as one of the subsequences (1) $e_i \ s \ e_j$ or (2) $s \ e_j \ e_i$ or (3) $e_j, \ e_i$, where $s$ is one of the signal words/phrases ``if'', ``whenever'', ``as long as'', ``on [the] condition that'', ``unless'', or ``provided that'', then we first remove all words in $e_i$ that are within the span of $e_j$, and, second, remove any signal word/phrase that is antecedent and adjacent to $e_j$.
In the few cases when no signal word or phrase was detected in the second step, we manually checked and revised the containing event.

\subsection{Analysis}

Table~\ref{tab-anno-info} shows the distribution of event pairs classified under each label in the annotated dataset, as well as the number of event pairs in the training, development and test document subsets.

It is not surprising that the distribution is biased, with a majority (just over 75\%) of event pairs labelled as ``argument''. The predicate-within-containing-event-span relation between events in the pairs we selected for annotation is a necessary condition for them to dependent.
Of the 2901 pairs labelled as non-argument (condition or independent), 2490 distinct events' representations are refined by the annotators and by our automated method.


\begin{table}[!htbp]
\small
\begin{center}
\tabcolsep=0.05cm
\begin{tabular}{l|rr|rr|rr} 
\hline
\textbf{} & \multicolumn{2}{c|}{\textbf{argument}}
& \multicolumn{2}{c|}{\textbf{non-argument}}
& \multicolumn{2}{c}{\textbf{overall}} \\

\hline
\textbf{} & \textbf{required} & \textbf{optional}
& \textbf{cond.} & \textbf{indep.} &  \\
\hline
\textbf{train}      & 4096 & 2837 & 335 & 1861 & 9129 & (77\%) \\
\textbf{dev}        & 635 & 421 & 41 & 355 & 1452 & (12.3\%) \\
\textbf{test}       & 594 & 368 & 70 & 239 & 1271 & (10.7\%) \\
\hline
\textbf{overall}    & 5325   & 3626   & 446   & 2455 & 11852 \\
                    & (44.9\%) & (30.6\%) & (3.8\%) & (20.7\%) & \\
\hline

\end{tabular}
\caption{Distribution of event pairs in the annotated EDeR dataset across labels and across the training, development and test subsets.}
\label{tab-anno-info}
\end{center} 
\end{table}

 
For human-annotated datasets, there is always a trade-off between the number of instances being annotated and the quality of annotations~\cite{kryscinski2019neural, cui2020mutual}. The size of our dataset is constrained by the annotation method. But it is comparable with or larger than many other human-annotated event relations reasoning datasets~\cite{pustejovsky2003timebank,glavas-etal-2014-hieve,chambers2014dense,mirza-tonelli-2016-catena,Ning:ACL:2018}.
A random sample of 5\% of the final set of annotations has been verified, and of these, at least 95\% were found to be correct.
This gives us confidence that the multi-level iterative annotation procedure produces high-quality annotation results.





\section{Experiment 1: Event Dependency Relation Extraction}
We first consider the task as binary classification. Assume that given a sentence $X$ including two events $e_i$ and $e_j$, the model is required to determine whether $e_j$ is an argument event of $e_i$ or not. We apply typical classification evaluation criteria including precision, recall, and accuracy. As with many existing studies~\cite{gunasekara-nejadgholi-2018-review,bars-sigir-2022,chen-etal-2022-meta}, we also report the results on ROC\_AUC~\cite{bradley1997rocauc}. It is used to measure the probability that a randomly chosen positive (i.e., argument) sample is ranked higher than a randomly chosen negative (i.e., non-argument) sample. A higher ROC\_AUC means better prediction performance.



\begin{table*}[!htbp]
\small
\tabcolsep=0.1cm

\begin{center}
\begin{tabular}{l|l|c|c|c|c}
\hline

\textbf{Input} & \textbf{Model}& \textbf{Precision (\%)} & \textbf{Recall (\%)} & \textbf{Accuracy(\%)} &\textbf{ROC\_AUC (\%)}\\

\hline
Sentence+predicates & Heuristic rules-based & \ \ \ \ \ \ \textbf{97.67 (1)}& 34.93& 50.08 &66.16\\
\hline

\multirow[t]{6}{*}{Event-Event Span}
& DistilBERT~\cite{sanh2019distilbert} &82.75 &90.23 &78.35 & 65.73\\
& BERT~\cite{devlin-etal-2019-bert} &87.53 &85.34 &79.69 & 70.61\\
& RoBERTa~\cite{liu2019roberta} &84.71 &87.53 &78.58 &69.09\\
& XLNet~\cite{yang2019xlnet} &83.61 & 89.60&78.82 &62.38 \\
& GPT-2~\cite{radford2019language} &85.91 &86.80 &79.21 &69.16 \\

\hline

\multirow[t]{6}{*}{Event-Event-SRL}
& DistilBERT~\cite{sanh2019distilbert} &84.75 &87.84 &78.82 &67.45 \\
& BERT~\cite{devlin-etal-2019-bert} &85.96 &87.21 &79.53 &71.37 \\
& RoBERTa~\cite{liu2019roberta} &82.89 & \ \ \ \ \ \ \textbf{91.16 (3)}&80.06 &72.71 \\
& XLNet~\cite{yang2019xlnet} &82.00 & \ \ \ \ \ \ \textbf{93.76 (1)} &79.69 &64.74 \\
& GPT-2~\cite{radford2019language} &86.60 &85.97 &79.29 & 72.20\\

\hline

\multirow[t]{6}{*}{Event-Event-SRL-DEP}
& DistilBERT~\cite{sanh2019distilbert} &85.48 &87.53 &79.29 &69.09 \\
& BERT~\cite{devlin-etal-2019-bert} &85.26 &89.60 &80.39&72.61\\
& RoBERTa~\cite{liu2019roberta} &83.21 & \ \ \ \ \ \ \textbf{91.16 (3)} &80.37 &71.18 \\
& XLNet~\cite{yang2019xlnet} &83.11 &91.06&79.21 &67.37 \\
& GPT-2~\cite{radford2019language} &85.89 &87.94 &79.92 &71.41 \\

\hline

\multirow[t]{6}{*}{Marked-predicate Sentence}
& DistilBERT~\cite{sanh2019distilbert} &82.13 & \ \ \ \ \ \ \textbf{93.14 (2)} & 79.45&71.72 \\
& BERT~\cite{devlin-etal-2019-bert} & \ \ \ \ \ \ \textbf{91.30 (2)} &85.14 & \ \ \ \ \ \ \textbf{82.61 (1)}& \ \ \ \ \ \ \textbf{74.69 (3)}\\
& RoBERTa~\cite{liu2019roberta} & \ \ \ \ \ \ \textbf{89.87 (3)} &85.85 & \ \ \ \ \ \ \textbf{81.97 (2)}& \ \ \ \ \ \ \textbf{75.40 (1)}\\
& XLNet~\cite{yang2019xlnet} &88.22 &86.38 & \ \ \ \ \ \ \textbf{80.94 (3)} & \ \ \ \ \ \ \textbf{75.17 (2)} \\
& GPT-2~\cite{radford2019language} &85.26 &89.60 &80.39 &70.61 \\

\hline

\end{tabular}
\caption{Comparison of the performance on the test set based on varying method and input combinations for the event dependency relation extraction task. The top-3 best results in each column are highlighted.}
\label{tab-result-task1}
\end{center} 
\end{table*}

\subsection{Baseline Models}
We evaluate the heuristic rules-based method, and several pre-trained language models, including discriminative models BERT~\cite{devlin-etal-2019-bert}, RoBERTa~\cite{liu2019roberta}, DistilBERT~\cite{sanh2019distilbert}, XLNet~\cite{yang2019xlnet}; as well as one autoregressive generative model GPT-2~\cite{radford2019language}.

\textbf{Heuristic rules-based method.} We design an unsupervised method for detecting if two events have a dependency relation. The rules are as follows: If any of them is satisfied, a contained event $e_j$ is an argument of the containing event $e_i$: 
(i) The syntactic dependency relation from the predicate of $e_i$ to the predicate of $e_j$ is clausal complement (\textit{ccomp} or \textit{xcomp}) or clausal subject (\textit{csubj}).
(ii) The syntactic dependency relation from the predicate of $e_j$ to the predicate of $e_i$ is copula (\textit{cop}).
(iii) All of $e_j$ is contained in an argument of $e_i$ that is labelled
with either ARGM-PRP (``purpose'') or ARGM-PNC (``purpose not cause'').
The syntactic dependency relations are obtained by dependency parsing using the Stanford CoreNLP toolkit~\cite{manning:ACL:2014}.

\subsection{Input Variations}
Besides various baseline models, we also explore the influence of different types of inputs. 

\textbf{Marked-Predicate Sentence Input} Inspired by the success of MarkedBERT~\cite{boualili2020markedbert} in information retrieval research, we add special marks to emphasize the predicate tokens in the sentence. For example, given a sentence ``The man who works here tells me to get the hell out.'' and two event predicates ``tells'' and ``get'', the marked-predicate input is ``The man who works here [V1] tells [\textbackslash V1] me to [V2] get [\textbackslash V2] the hell out.''. 

\textbf{Event-Event Span Input} We also create the two events' span style input. For the same example as above, the Event-Event Span style input is ```The man who works here \{V: tells\} me to get the hell out. [SEP] me \{V: get\} the hell out''. In which, we add a special token ``[SEP]'' for separation.

\textbf{Event-Event-SRL Input} For further checking the influence of the SRL labels in helping classify the Event-Event relations, we add the argument label of the argument where the contained event predicate is in the containing event. We also add a special token ``[SRL]'' for model recognition.

\textbf{Event-Event-SRL-DEP Input} Based on the Event-Event-SRL Input, we also add the syntactic dependency relation information of the two event predicates. A special token ``[DEP]'' is specified for model recognition.

\subsection{Implementation Settings}
For the fairness of comparison, all the models are trained on a computer with 64 CPUs with 126GB of RAM and one RTX-3090 GPU with 24GB of RAM. 
We set the optimization method as AdamW~\cite{loshchilov2019adamw} which is adopted with an initial learning rate of $1\mathrm{e}{-5}$ and a batch size of 4 and a number of epochs of 4 during the finetuning process of each model. The model with the highest F1-score on the development set is selected.

\subsection{Results and Analysis}

The models' performances using various inputs are shown in Table~\ref{tab-result-task1}. We observe that compared with only using the Event-Event span as input, adding the event predicate's syntactic dependency and the (semantic) argument label information improves the performance of most of the models. However, using the marked-predicate single sentence as input is more effective than these event-event style inputs. 
Taking the marked-predicate sentence as input, the BERT and RoBERTa models outperform others in terms of accuracy (82.61\% and 81.97\%), precision (91.30\% and 89.87\%), and ROC\_AUC (74.69\% and 75.40\%). 
GPT-2 performs undesirably compared with the discriminative models, indicating that this generative model still has room to improve its reasoning ability. Specifically, although the recall of the heuristic rules-based method is 34.93\% and much lower than others, it achieves the highest precision at 97.67\%. It indicates the latent correlations between our event dependency and predicates' syntactic/semantic dependency relations. As an unsupervised method, it is helpful for precisely predicting the existence of the dependency relation between two unseen events. The highest accuracy of the baseline models reaches 82.61\%. It indicates our EDeR dataset is sufficient for training a good predictor for recognizing the event dependency relations based on the language model's pre-train\&fine-tune mechanism.

 \begin{figure}[!t]
\centering
\includegraphics[width=0.48\textwidth]{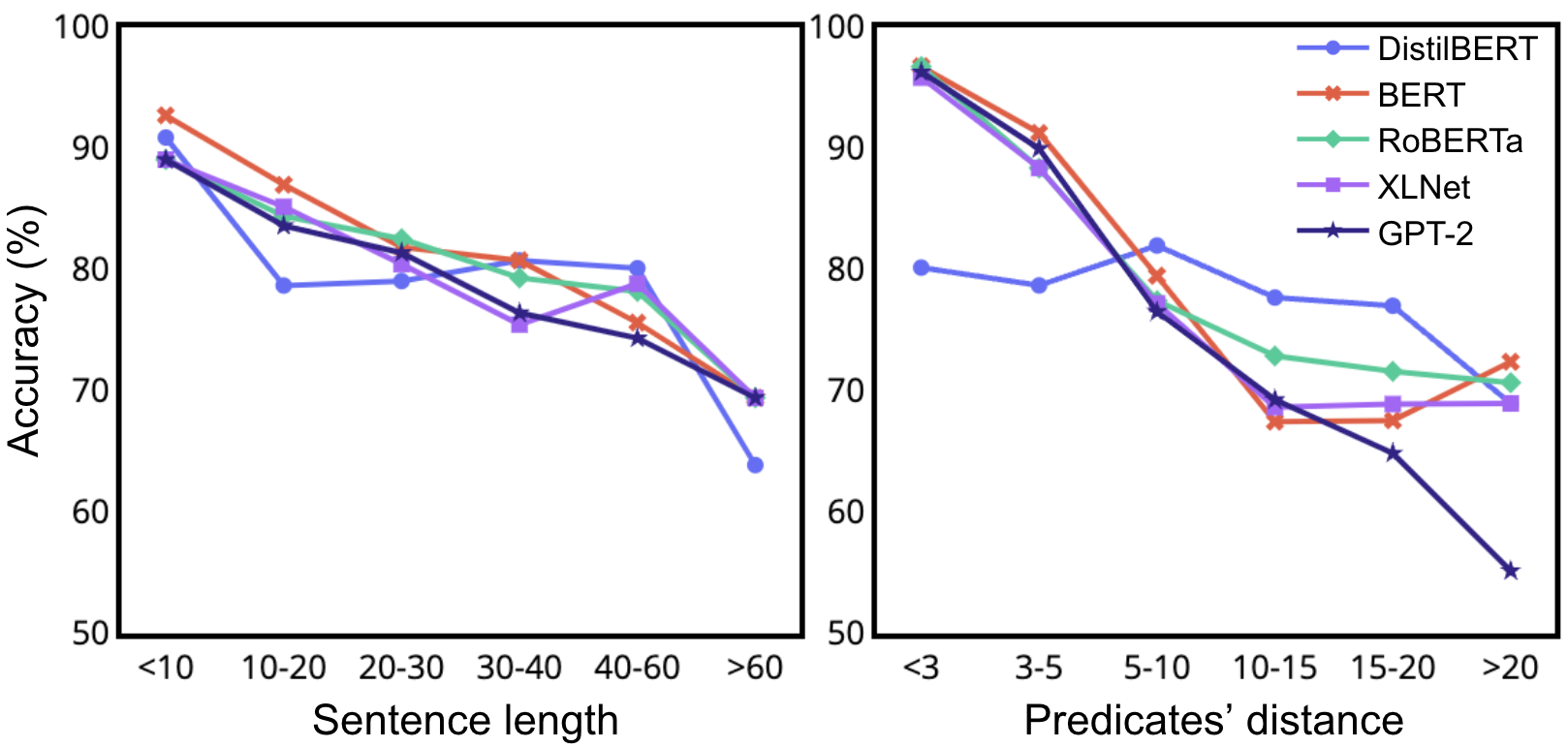}
\caption{Baseline model performances (accuracy) across different ranges of sentence length (Left) and different ranges of predicates' distance (Right).}
\label{fig-impact-exp1}
\end{figure}

\subsubsection{Impact of Sentence Length and Predicate Distance}
Figure~\ref{fig-impact-exp1} (left) illustrates the performance of the baseline models across sentences with different lengths, given the input as the marked-predicate sentences. As the sentence length increases, the performance of almost all the models decreases accordingly, which is consistent with human intuition. Longer sentences may contain more complex structures, which are challenging for models to capture. Similarly, as shown in Figure~\ref{fig-impact-exp1} (right), the models' accuracy also decreases when the distance between two predicates (i.e., the number of words between them) in the sentence increases. It also indicates that detecting the dependency relations between distant predicates is challenging for the models.

\section{Experiment 2: Benefit Downstream Tasks}
In this section, we show that EDeR can be used to improve the performance of downstream tasks.

\subsection{Guide Event Representation Extraction}
Event representation extraction is performed as a semantic role labelling (SRL) task that identifies arguments (including spans and labels) of the event, given the event's predicates and its sentence. 
 Although originally based on the OntoNotes semantic role labels, our revised task is based on the updated semantic role labels from EDeR. 
 
\begin{figure}[!t]
\centering
\includegraphics[width=0.48\textwidth]{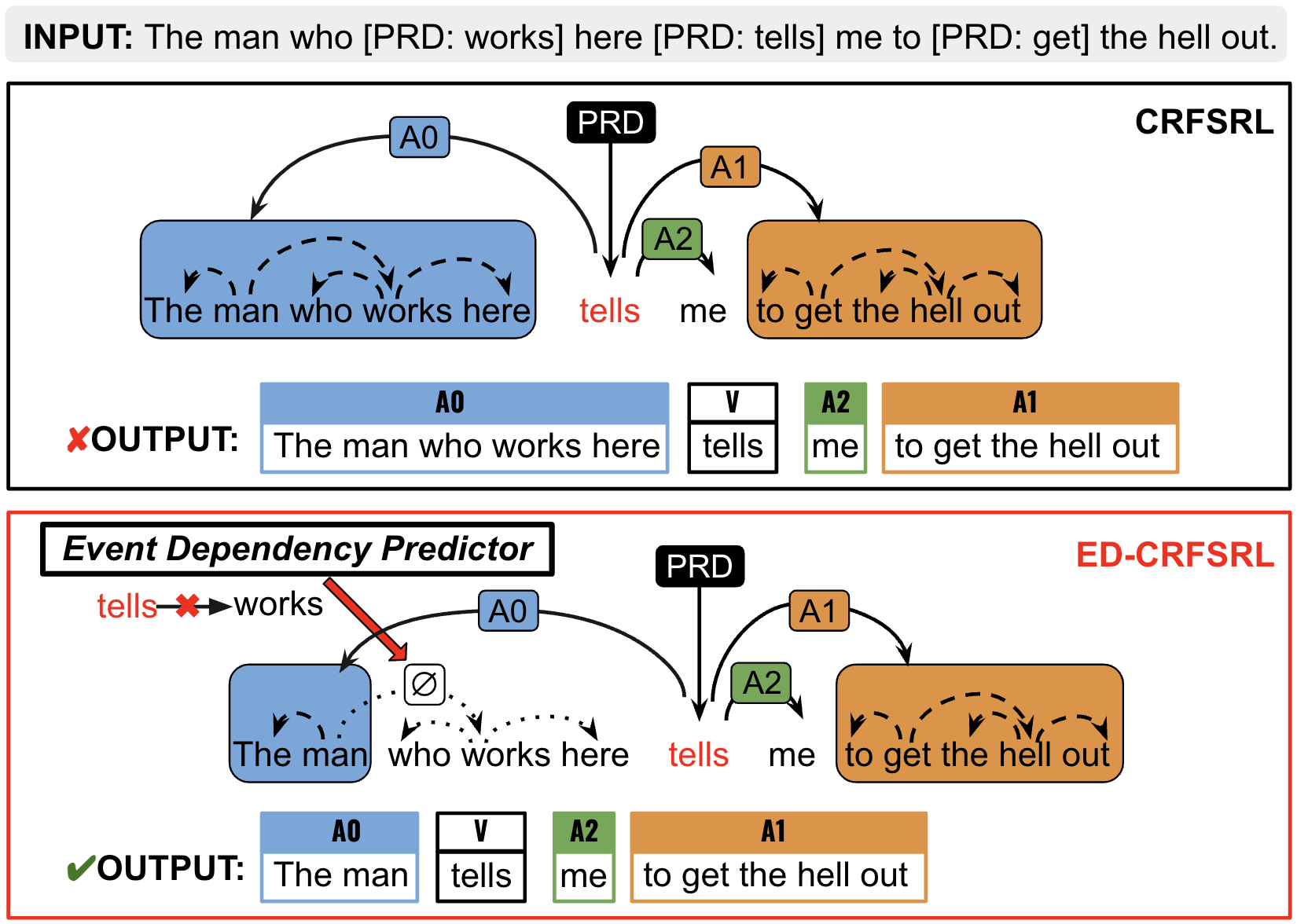}
\caption{An example of different event representation outputs from the CRFSRL system~\protect\cite{zhang-etal-2022-crfsrl} and our ED-CRFSRL system.}
\label{fig-srl-case-study}
\end{figure}

In practice, we select the SOTA SRL system CRFSRL ~\cite{zhang-etal-2022-crfsrl} trained on OntoNotes as the baseline.
As Figure~\ref{fig-srl-case-study} CRFSRL part shows, the system takes the input of a sentence with marked event predicate ($p$) and detects all the words that have a syntactic dependency with $p$ as descendants. Then it builds latent sub-trees for each descendant based on the syntactic dependencies between the descendant and other words within the sentence. The sub-trees are the argument spans. 

In Figure~\ref{fig-srl-case-study} CRFSRL example, its event representation output contains a non-argument event ``The man who works here''. Therefore, we modify the CRFSRL by integrating event dependency information with it. The modified system is named Event Dependency CRFSRL (ED-CRFSRL). As Figure~\ref{fig-srl-case-study} ED-CRFSRL part shows, based on the event dependency relation prediction, event (predicate) ``works'' is not an argument of ``tells'', all the sub-trees that root at ``works'' are then pruned. So the correct event representation ``The man tells me to get the hell out'' is produced by our ED-CRFSRL system.


We evaluate the performances of the SRL systems based on the official scripts provided by CoNLL-2012 shared task\footnote{https://www.cs.upc.edu/~srlconll/}. 
We initially investigate the performance of the CRFSRL system using the original event representations from the (275 sampled documents) subset of OntoNotes. The system is named CRFSRL-O. The CRFSRL-O system's testing precision, recall and F1-score are 77.07\%, 81.03\% and 79.00\%, respectively. On the basis of all the updated events from EDeR, we train the CRFSRL system. As Table~\ref{tab-srl-perform} first line shows, CRFSRL achieves 74.19\% precision, 78.45\% recall and 76.26\% F1-score, which are overall lower than CRFSRL-O's performance. It indicates the updated event representations in EDeR are more difficult to model for the CRFSRL system. However, the performance of ED-CRFSRL (P) which uses the baseline model (typically the highest-accuracy baseline model from Experiment 1) predicted event dependency relations and the performance of ED-CRFSRL (G) which uses the EDeR annotated event dependency information are both better than CRFSRL. Especially for ED-CRFSRL (G), it achieves 76.09\% precision, 78.81\% recall and 77.43\% F1-score.

Specifically, the performance gap is more significant for the refined events only. The CRFSRL's performance is inferior to its performance on all the updated events, which indicates it cannot deal with the refined events as well as other non-refined events. Guided by either the baseline model's predicted or human-annotated event dependency relation information, the ED-CRFSRL's performance gets much improved (e.g., 13.8\% and 20.77\% increase in precision, and 7.44\% and 11.18\% increase in F1-score), compared with the CRFSRL model that does not use this information. 

All of the results prove that the event dependency information is essential to improve the SRL system for the revised event representation extraction task, especially for extracting the refined events.

\begin{table}[!htbp]
\small
\tabcolsep=0.1cm
\setlength{\arrayrulewidth}{1.5pt}
\begin{center}
\begin{tabular}{lccc} 
\textbf{} & \textbf{Precision} & \textbf{Recall} & \textbf{F1-score} \\
\hline
\multicolumn{4}{l}{\textbf{All updated events}} \\
\hline
CRFSRL~\cite{zhang-etal-2022-crfsrl}      & 74.19 & 78.45 & 76.26\\
ED-CRFSRL (P)    & 74.35 & 78.53& 76.38\\
ED-CRFSRL (G)   & \textbf{76.09} & \textbf{78.81} &\textbf{77.43}\\

\hline
\multicolumn{4}{l}{\textbf{Refined events only}} \\
\hline
CRFSRL~\cite{zhang-etal-2022-crfsrl}    & 67.34 & 65.73 & 66.53\\
ED-CRFSRL (P)  & 81.14 &67.97 &73.97\\
ED-CRFSRL (G)  & \textbf{88.11} & \textbf{69.51} & \textbf{77.71}\\

\hline
\end{tabular}
\caption{For the revised event extraction task, the performance comparison between the CRFSRL system \protect\cite{zhang-etal-2022-crfsrl}, the ED-CRFSRL system using baseline model predicted (ED-CRFSRL (P)) and EDeR annotated (ED-CRFSRL (G)) event dependency information. \textbf{All updated events} refer to the performance on all the events (including refined and non-refined) from the EDeR test set, while \textbf{Refined events only} means the performance on the refined event representations only.}
\label{tab-srl-perform}
\end{center} 
\end{table}

\subsection{Improve Co-reference Resolution}

To further investigate whether the refined event representations from EDeR are more reasonable than the original version provided by OntoNotes, we apply them to a downstream task: co-reference resolution. 

\begin{figure}[!t]
\centering
\includegraphics[width=0.48\textwidth]{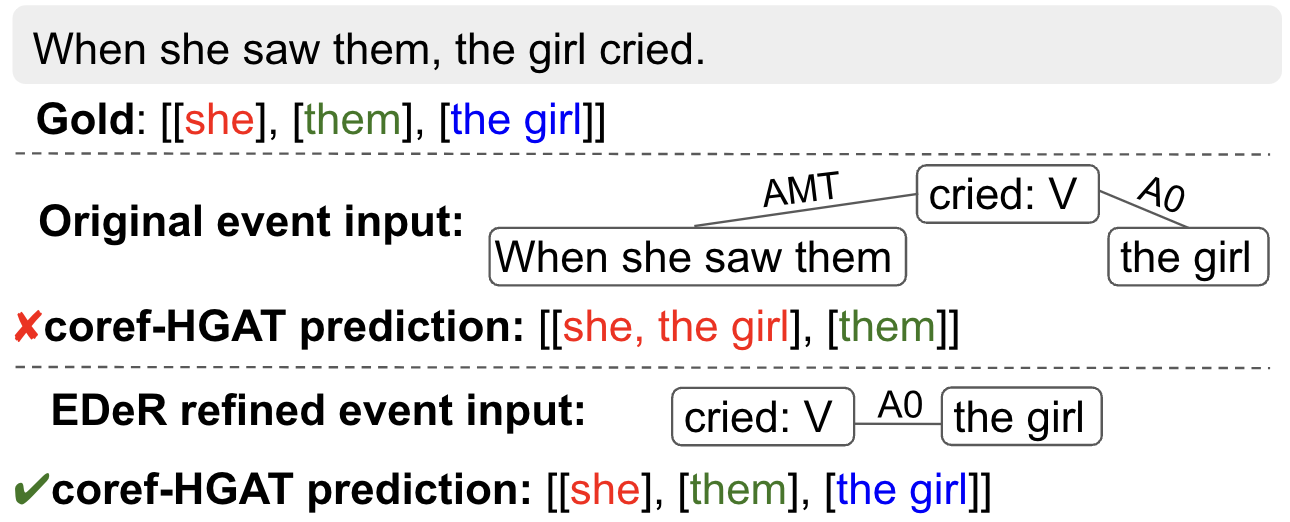}
\caption{An example of different co-reference resolution outputs of the coref-HGAT model taking the original event representation from OntoNotes and our EDeR refined event representation as inputs.}
\label{fig-coref-case-study}
\end{figure}

Given a text, co-reference resolution (CR) aims to identify all mentions that refer to the same entity. We select a SOTA CR model called coref-HGAT~\cite{jiang-cohn-2021-coref-hgat}, which is initially trained and tested based on the OntoNotes CR annotations. 
The coref-HGAT model encodes both event semantic role information and word syntactic dependency information in heterogeneous graphs, and incorporates the graphs into contextualized embeddings to identify the co-reference links between mentions.  
As Figure~\ref{fig-coref-case-study} example shows, in the original event representation (from OntoNotes), the event ``When she saw them'' is wrongly linked as an argument of the event ``cried'' in the graph. The coref-HGAT model wrongly predicts that the mention ``the girl'' co-refers to ``she''. But based on the refined event representation from EDeR, there is no semantic link between ``she'' and ``the girl'' in the (updated) graph. The model correctly detects that the mentions ``she'' and ``the girl'' actually do not co-refer to one another.

 \begin{table}[!htbp]
 \small
\tabcolsep=0.1cm
\setlength{\arrayrulewidth}{1.5pt}
\begin{center}
\begin{tabular}{lccc} 
\textbf{Input} & \textbf{Precision} & \textbf{Recall} & \textbf{F1-score} \\
\hline
OntoNotes Event Reps. (G)      & 67.73 & 65.41 & 66.50\\
EDeR Event Reps. (G)   & \textbf{69.14} & \textbf{65.69} & \textbf{67.40}\\
\hline
OntoNotes Event Reps. (P)       &64.82  & 63.79 & 64.28\\
EDeR Event Reps. (P)     & \textbf{67.79} & \textbf{64.42} & \textbf{65.96}\\
\hline

\end{tabular}
\caption{Comparison of the co-reference resolution performance of the coref-HGAT~\protect\cite{jiang-cohn-2021-coref-hgat} model using the annotated (G) and the predicted (P) original event representations from OntoNotes and updated event representations from EDeR.}
\label{tab-coref-perform}
\end{center} 
\end{table}
 As EDeR is based on a subset (275 sampled documents) of OntoNotes, we conduct this co-reference resolution task using the CF annotations from this subset. The evaluation is still based on the official CoNLL-2012 evaluation scripts.

First, when using the EDeR annotated updated event representations as input for the coref-HGAT model, its testing precision, recall, and F1-score are increased by 1.41\%, 0.28\%, and 0.9\%, respectively, compared to the original annotated event representations (from OntoNotes's 275 sampled documents), as shown in the first two lines of Table~\ref{tab-coref-perform}. The results prove that the EDeR's refined event representations are more beneficial to the co-reference resolution task, which further proves the plausibility and validity of our event representation revision.

In addition, we also examine the predictive performance of the coref-HGAT model using the CRFSRL-O predicted original event representations and the ED-CRFSRL (P) predicted updated event representations from Section 5.1.
As Table~\ref{tab-coref-perform} last two lines show, the precision, recall and F1-score of the coref-HGAT model using the predicted updated event representations are all higher than using the predicted original event representations by 2.97\%, 0.63\% and 1.68\%, respectively. Although containing model prediction errors, the coref-HGAT model's performance using the predicted updated event representations is comparable to the results obtained using the annotated OntoNotes event representations.
The results show the possibility of applying the co-reference resolution model to a more general scenario by replacing its annotated original event representation input with our predictable updated event representations.


\begin{table*}[!htbp]
\small
\tabcolsep=0.1cm
\begin{center}
\begin{tabular}{l|l|c|c|c|c}
\hline

\textbf{Input} & \textbf{Model} & \textbf{Precision (\%)} & \textbf{Recall(\%)}& \textbf{F1-score(\%)}  & \textbf{Accuracy(\%)} \\
\hline

\multirow[t]{5}{*}{Event-Event Span}
& DistilBERT~\cite{sanh2019distilbert} &61.05 &60.97 &60.90 &64.80 \\
& BERT~\cite{devlin-etal-2019-bert}  &65.33 &65.64 &64.92 &69.13 \\
& RoBERTa~\cite{liu2019roberta}  &65.63 &65.56 &65.59 &69.06 \\
& XLNet~\cite{yang2019xlnet}  &62.52 &62.46 &62.47 &66.46 \\
& GPT-2~\cite{radford2019language} &61.70 &61.49 &61.49 &65.43 \\

\hline

\multirow[t]{5}{*}{Event-Event-SRL}
& DistilBERT~\cite{sanh2019distilbert} &63.74 &60.90 &61.41 &65.98 \\
& BERT~\cite{devlin-etal-2019-bert}  &65.68 &65.89 &65.77 &69.21 \\
& RoBERTa~\cite{liu2019roberta}  & \ \ \ \ \ \ \textbf{66.08 (3)}  &66.44 &66.08 & \ \ \ \ \ \ \textbf{69.69 (3)}  \\
& XLNet~\cite{yang2019xlnet}  &65.71 &66.04 &65.28 &69.21 \\
& GPT-2~\cite{radford2019language} &63.04 &62.70 &62.73 &66.46 \\

\hline

\multirow[t]{5}{*}{Event-Event-SRL-DEP}
& DistilBERT~\cite{sanh2019distilbert} &63.81 &62.82 &63.15 &66.77 \\
& BERT~\cite{devlin-etal-2019-bert}  & \ \ \ \ \ \ \textbf{66.21 (2)}  &66.13 & \ \ \ \ \ \ \textbf{66.12 (3)} & \ \ \ \ \ \ \textbf{69.76 (2)} \\
& RoBERTa~\cite{liu2019roberta}  & \ \ \ \ \ \ \textbf{67.21 (1)} & \ \ \ \ \ \ \textbf{67.31 (1)} & \ \ \ \ \ \ \textbf{67.15 (1)} & \ \ \ \ \ \ \textbf{70.79 (1)} \\
& XLNet~\cite{yang2019xlnet}  &66.01 & \ \ \ \ \ \ \textbf{66.21 (3)} &66.08 &69.37 \\
& GPT-2~\cite{radford2019language} &62.86 &62.99 &62.82 &66.54 \\

\hline

\multirow[t]{5}{*}{Marked-predicate Sentence}
& DistilBERT~\cite{sanh2019distilbert} &62.18 &60.99 &61.40 &64.96 \\
& BERT~\cite{devlin-etal-2019-bert}  &65.91 & \ \ \ \ \ \ \textbf{66.67 (2)} & \ \ \ \ \ \ \textbf{66.13 (2)} &69.29 \\
& RoBERTa~\cite{liu2019roberta}  &65.83 &66.17 &65.89 &69.53 \\
& XLNet~\cite{yang2019xlnet}  &64.83 &65.69 &64.78 &68.11 \\
& GPT-2~\cite{radford2019language} &60.70 &60.88 &60.74 &64.33 \\

\hline

\end{tabular}
\caption{Comparison of the performance for the fine-grained event dependency relation classification task. The top-3 best results in each column are highlighted. Precision, recall and F1-score results are all macro-averaged.}
\label{tab-result-task2}
\end{center} 
\end{table*}

\section{Experiment 3: Fine-grained Event Dependency Relation Extraction}
As described, for our EDeR dataset, the argument label can be further categorized into two fine-grained classes: \textit{required argument} and \textit{optional argument}. We apply the introduced baseline models and inputs to this three-way classification task. 

 \begin{figure}[!t]
\centering
\includegraphics[width=0.48\textwidth]{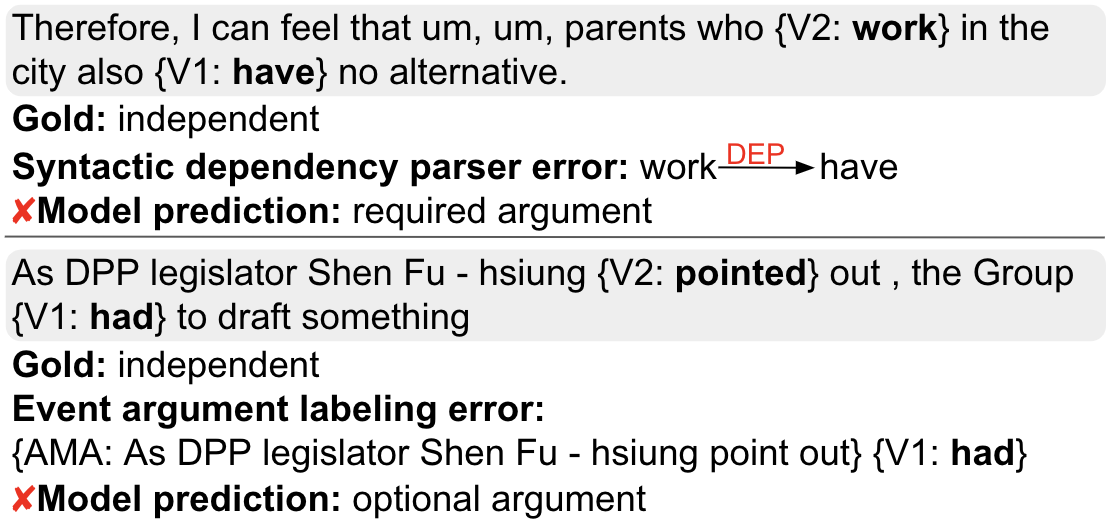}
\caption{Case study for fine-grained event dependency relation extraction.}
\label{fig-multi-classification-case-study}
\end{figure}

The prediction results of the baseline language models taking various types of inputs are shown in Table~\ref{tab-result-task2}. First, the Event-Event-SRL-DEP input achieves overall better performance than others in this task. Besides the useful information provided by the syntactic dependencies as discussed in Section 4.4, another possible reason is that in the EDeR dataset, the predicates of over 95.79\% of the events labelled as required arguments are located within the numbered arguments (ARG0--ARG5) of their containing events. Second, similar to Experiment 1 results, still the BERT and RoBERTa models outperform. Specifically, RoBERTa achieves the best precision, recall, F1-score and accuracy results, which reveals its better relation reasoning capacity. Third, compared with the binary classification on argument/non-argument, the overall performance is dropped - the highest accuracy is 70.79\%. It demonstrates that the three-way classification is more challenging for our current baseline models.




\subsection{Case Study} 
We identify two phenomena from the Event-Event-SRL-DEP based RoBERTa model for the fine-grained event dependency relation extraction:

\subsubsection{Syntactic Dependency Parsing Errors} Some incorrect classification results may be because of the misleading syntactic dependency parsing results. As shown in the first case of Figure~\ref{fig-multi-classification-case-study}, the off-the-shelf dependency parser wrongly assigns syntactic dependency relation between the two event predicates ``work'' and ``have'', which could affect the model prediction.

\subsubsection{Event Argument Labelling Errors} Furthermore, as the second case of Figure~\ref{fig-multi-classification-case-study} shows, the semantic labels provided by OntoNotes for the event predicate ``had'' is incorrect (e.g., ``the group'' is missing as the ARG0). It may prevent the model from accurately predicting the event dependency relation types.

\section{Conclusion}
We introduced EDeR, a high-quality human-annotated event dependency relation dataset. We presented a thorough description of the dataset construction process and an analysis of it. The experimental results of various competitive baselines, as well as subsequent applications on downstream tasks, demonstrate the utility and benefit of the dataset. Further baseline performance on predicting the three-way classification into the required, optional and non-argument demonstrates the challenge of the dataset. It will direct our future work, and we hope it will encourage future research in the IR and NLP communities as well. We open-source the EDeR dataset and the code of related baseline models.



\bibliography{anthology,custom}

\begin{thebibliography}{55}
\expandafter\ifx\csname natexlab\endcsname\relax\def\natexlab#1{#1}\fi

\bibitem[{Araki et~al.(2014)Araki, Liu, Hovy, and
  Mitamura}]{araki2014detecting}
Jun Araki, Zhengzhong Liu, Eduard Hovy, and Teruko Mitamura. 2014.
\newblock \href
  {http://www.lrec-conf.org/proceedings/lrec2014/pdf/963_Paper.pdf} {Detecting
  subevent structure for event coreference resolution}.
\newblock In \emph{Proceedings of the Ninth International Conference on
  Language Resources and Evaluation ({LREC}'14)}, pages 4553--4558, Reykjavik,
  Iceland. European Language Resources Association (ELRA).

\bibitem[{Balazs et~al.(2017)Balazs, Marrese-Taylor, Loyola, and
  Matsuo}]{balazs-etal-2017-refining}
Jorge Balazs, Edison Marrese-Taylor, Pablo Loyola, and Yutaka Matsuo. 2017.
\newblock \href {https://doi.org/10.18653/v1/W17-5310} {Refining raw sentence
  representations for textual entailment recognition via attention}.
\newblock In \emph{Proceedings of the 2nd Workshop on Evaluating Vector Space
  Representations for {NLP}}, pages 51--55, Copenhagen, Denmark. Association
  for Computational Linguistics.

\bibitem[{Bethard(2013)}]{bethard:SemEval:2013}
Steven Bethard. 2013.
\newblock \href {https://aclanthology.org/S13-2002} {{C}lear{TK}-{T}ime{ML}: A
  minimalist approach to {T}emp{E}val 2013}.
\newblock In \emph{Second Joint Conference on Lexical and Computational
  Semantics (*{SEM}), Volume 2: Proceedings of the Seventh International
  Workshop on Semantic Evaluation ({S}em{E}val 2013)}, pages 10--14, Atlanta,
  Georgia, USA. Association for Computational Linguistics.

\bibitem[{Bonial et~al.(2012)Bonial, Hwang, Bonn, Conger, Babko-Malaya, and
  Palmer}]{bonial:journal:2012}
Claire Bonial, Jena Hwang, Julia Bonn, Kathryn Conger, Olga Babko-Malaya, and
  Martha Palmer. 2012.
\newblock English propbank annotation guidelines.
\newblock \emph{Center for Computational Language and Education Research
  Institute of Cognitive Science University of Colorado at Boulder}, 48.

\bibitem[{Boualili et~al.(2020)Boualili, Moreno, and
  Boughanem}]{boualili2020markedbert}
Lila Boualili, Jose~G. Moreno, and Mohand Boughanem. 2020.
\newblock \href {https://doi.org/10.1145/3397271.3401194} {Markedbert:
  Integrating traditional ir cues in pre-trained language models for passage
  retrieval}.
\newblock In \emph{Proceedings of the 43rd International ACM SIGIR Conference
  on Research and Development in Information Retrieval}, SIGIR '20, page
  1977–1980, New York, NY, USA. Association for Computing Machinery.

\bibitem[{Bradley(1997)}]{bradley1997rocauc}
Andrew~P Bradley. 1997.
\newblock The use of the area under the roc curve in the evaluation of machine
  learning algorithms.
\newblock \emph{Pattern recognition}, 30(7):1145--1159.

\bibitem[{Carreras and M{\`a}rquez(2005)}]{carreras-marquez-2005-conll}
Xavier Carreras and Llu{\'\i}s M{\`a}rquez. 2005.
\newblock \href {https://aclanthology.org/W05-0620} {Introduction to the
  {C}o{NLL}-2005 shared task: Semantic role labeling}.
\newblock In \emph{Proceedings of the Ninth Conference on Computational Natural
  Language Learning ({C}o{NLL}-2005)}, pages 152--164, Ann Arbor, Michigan.
  Association for Computational Linguistics.

\bibitem[{Cassidy et~al.(2014)Cassidy, McDowell, Chambers, and
  Bethard}]{cassidy2014annotation}
Taylor Cassidy, Bill McDowell, Nathanel Chambers, and Steven Bethard. 2014.
\newblock An annotation framework for dense event ordering.
\newblock Technical report, Carnegie-Mellon Univ Pittsburgh PA.

\bibitem[{Chambers(2013)}]{chambers:2013}
Nathanael Chambers. 2013.
\newblock Navytime: Event and time ordering from raw text.
\newblock Technical report, Naval Academy, Annapolis MD.

\bibitem[{Chambers et~al.(2014{\natexlab{a}})Chambers, Cassidy, McDowell, and
  Bethard}]{chambers:ACL:2014}
Nathanael Chambers, Taylor Cassidy, Bill McDowell, and Steven Bethard.
  2014{\natexlab{a}}.
\newblock Dense event ordering with a multi-pass architecture.
\newblock \emph{Transactions of the Association for Computational Linguistics},
  2:273--284.

\bibitem[{Chambers et~al.(2014{\natexlab{b}})Chambers, Cassidy, McDowell, and
  Bethard}]{chambers2014dense}
Nathanael Chambers, Taylor Cassidy, Bill McDowell, and Steven Bethard.
  2014{\natexlab{b}}.
\newblock Dense event ordering with a multi-pass architecture.
\newblock \emph{Transactions of the Association for Computational Linguistics},
  2:273--284.

\bibitem[{Chen et~al.(2020)Chen, Lan, Du, and Lobanov}]{chen-etal-2020-joint}
Miao Chen, Ganhui Lan, Fang Du, and Victor Lobanov. 2020.
\newblock \href {https://doi.org/10.18653/v1/2020.clinicalnlp-1.26} {Joint
  learning with pre-trained transformer on named entity recognition and
  relation extraction tasks for clinical analytics}.
\newblock In \emph{Proceedings of the 3rd Clinical Natural Language Processing
  Workshop}, pages 234--242, Online. Association for Computational Linguistics.

\bibitem[{Chen et~al.(2022)Chen, Zhong, Zha, Karypis, and
  He}]{chen-etal-2022-meta}
Yanda Chen, Ruiqi Zhong, Sheng Zha, George Karypis, and He~He. 2022.
\newblock \href {https://doi.org/10.18653/v1/2022.acl-long.53} {Meta-learning
  via language model in-context tuning}.
\newblock In \emph{Proceedings of the 60th Annual Meeting of the Association
  for Computational Linguistics (Volume 1: Long Papers)}, pages 719--730,
  Dublin, Ireland. Association for Computational Linguistics.

\bibitem[{Cui et~al.(2020)Cui, Wu, Liu, Zhang, and Zhou}]{cui2020mutual}
Leyang Cui, Yu~Wu, Shujie Liu, Yue Zhang, and Ming Zhou. 2020.
\newblock \href {https://doi.org/10.18653/v1/2020.acl-main.130} {{M}u{T}ual: A
  dataset for multi-turn dialogue reasoning}.
\newblock In \emph{Proceedings of the 58th Annual Meeting of the Association
  for Computational Linguistics}, pages 1406--1416, Online. Association for
  Computational Linguistics.

\bibitem[{Devlin et~al.(2019)Devlin, Chang, Lee, and
  Toutanova}]{devlin-etal-2019-bert}
Jacob Devlin, Ming-Wei Chang, Kenton Lee, and Kristina Toutanova. 2019.
\newblock \href {https://doi.org/10.18653/v1/N19-1423} {{BERT}: Pre-training of
  deep bidirectional transformers for language understanding}.
\newblock In \emph{Proceedings of the 2019 Conference of the North {A}merican
  Chapter of the Association for Computational Linguistics: Human Language
  Technologies, Volume 1 (Long and Short Papers)}, pages 4171--4186,
  Minneapolis, Minnesota. Association for Computational Linguistics.

\bibitem[{Ethayarajh(2019)}]{ethayarajh-2019-contextual}
Kawin Ethayarajh. 2019.
\newblock \href {https://doi.org/10.18653/v1/D19-1006} {How contextual are
  contextualized word representations? {C}omparing the geometry of {BERT},
  {ELM}o, and {GPT}-2 embeddings}.
\newblock In \emph{Proceedings of the 2019 Conference on Empirical Methods in
  Natural Language Processing and the 9th International Joint Conference on
  Natural Language Processing (EMNLP-IJCNLP)}, pages 55--65, Hong Kong, China.
  Association for Computational Linguistics.

\bibitem[{Fischbach et~al.(2021)Fischbach, Frattini, Spaans, Kummeth,
  Vogelsang, Mendez, and Unterkalmsteiner}]{fischbach:etal:2021}
Jannik Fischbach, Julian Frattini, Arjen Spaans, Maximilian Kummeth, Andreas
  Vogelsang, Daniel Mendez, and Michael Unterkalmsteiner. 2021.
\newblock Automatic detection of causality in requirement artifacts: the cira
  approach.
\newblock In \emph{International Working Conference on Requirements
  Engineering: Foundation for Software Quality}, pages 19--36, Cham. Springer
  International Publishing.

\bibitem[{Glava{\v{s}} and
  {\v{S}}najder(2014)}]{glavas-snajder-2014-constructing}
Goran Glava{\v{s}} and Jan {\v{S}}najder. 2014.
\newblock \href {https://doi.org/10.3115/v1/W14-3705} {Constructing coherent
  event hierarchies from news stories}.
\newblock In \emph{Proceedings of {T}ext{G}raphs-9: the workshop on Graph-based
  Methods for Natural Language Processing}, pages 34--38, Doha, Qatar.
  Association for Computational Linguistics.

\bibitem[{Glava{\v{s}} et~al.(2014)Glava{\v{s}}, {\v{S}}najder, Moens, and
  Kordjamshidi}]{glavas-etal-2014-hieve}
Goran Glava{\v{s}}, Jan {\v{S}}najder, Marie-Francine Moens, and Parisa
  Kordjamshidi. 2014.
\newblock \href
  {http://www.lrec-conf.org/proceedings/lrec2014/pdf/1023_Paper.pdf}
  {{H}i{E}ve: A corpus for extracting event hierarchies from news stories}.
\newblock In \emph{Proceedings of the Ninth International Conference on
  Language Resources and Evaluation ({LREC}'14)}, pages 3678--3683, Reykjavik,
  Iceland. European Language Resources Association (ELRA).

\bibitem[{Gunasekara and Nejadgholi(2018)}]{gunasekara-nejadgholi-2018-review}
Isuru Gunasekara and Isar Nejadgholi. 2018.
\newblock \href {https://doi.org/10.18653/v1/W18-5103} {A review of standard
  text classification practices for multi-label toxicity identification of
  online content}.
\newblock In \emph{Proceedings of the 2nd Workshop on Abusive Language Online
  ({ALW}2)}, pages 21--25, Brussels, Belgium. Association for Computational
  Linguistics.

\bibitem[{Haji{\v{c}} et~al.(2009)Haji{\v{c}}, Ciaramita, Johansson, Kawahara,
  Mart{\'\i}, M{\`a}rquez, Meyers, Nivre, Pad{\'o}, {\v{S}}t{\v{e}}p{\'a}nek,
  Stra{\v{n}}{\'a}k, Surdeanu, Xue, and Zhang}]{hajic-etal-2009-conll}
Jan Haji{\v{c}}, Massimiliano Ciaramita, Richard Johansson, Daisuke Kawahara,
  Maria~Ant{\`o}nia Mart{\'\i}, Llu{\'\i}s M{\`a}rquez, Adam Meyers, Joakim
  Nivre, Sebastian Pad{\'o}, Jan {\v{S}}t{\v{e}}p{\'a}nek, Pavel
  Stra{\v{n}}{\'a}k, Mihai Surdeanu, Nianwen Xue, and Yi~Zhang. 2009.
\newblock \href {https://aclanthology.org/W09-1201} {The {C}o{NLL}-2009 shared
  task: Syntactic and semantic dependencies in multiple languages}.
\newblock In \emph{Proceedings of the Thirteenth Conference on Computational
  Natural Language Learning ({C}o{NLL} 2009): Shared Task}, pages 1--18,
  Boulder, Colorado. Association for Computational Linguistics.

\bibitem[{Hovy et~al.(2013)Hovy, Mitamura, Verdejo, Araki, and
  Philpot}]{hovy2013events}
Eduard Hovy, Teruko Mitamura, Felisa Verdejo, Jun Araki, and Andrew Philpot.
  2013.
\newblock Events are not simple: Identity, non-identity, and quasi-identity.
\newblock In \emph{Workshop on events: Definition, detection, coreference, and
  representation}, pages 21--28, Atlanta, Georgia. Association for
  Computational Linguistics.

\bibitem[{Huguet~Cabot and
  Navigli(2021)}]{huguet-cabot-navigli-2021-rebel-relation}
Pere-Llu{\'\i}s Huguet~Cabot and Roberto Navigli. 2021.
\newblock \href {https://doi.org/10.18653/v1/2021.findings-emnlp.204} {{REBEL}:
  Relation extraction by end-to-end language generation}.
\newblock In \emph{Findings of the Association for Computational Linguistics:
  EMNLP 2021}, pages 2370--2381, Punta Cana, Dominican Republic. Association
  for Computational Linguistics.

\bibitem[{Jiang and Cohn(2021)}]{jiang-cohn-2021-coref-hgat}
Fan Jiang and Trevor Cohn. 2021.
\newblock \href {https://doi.org/10.18653/v1/2021.naacl-main.125}
  {Incorporating syntax and semantics in coreference resolution with
  heterogeneous graph attention network}.
\newblock In \emph{Proceedings of the 2021 Conference of the North American
  Chapter of the Association for Computational Linguistics: Human Language
  Technologies}, pages 1584--1591, Online. Association for Computational
  Linguistics.

\bibitem[{Kryscinski et~al.(2019)Kryscinski, Keskar, McCann, Xiong, and
  Socher}]{kryscinski2019neural}
Wojciech Kryscinski, Nitish~Shirish Keskar, Bryan McCann, Caiming Xiong, and
  Richard Socher. 2019.
\newblock \href {https://doi.org/10.18653/v1/D19-1051} {Neural text
  summarization: A critical evaluation}.
\newblock In \emph{Proceedings of the 2019 Conference on Empirical Methods in
  Natural Language Processing and the 9th International Joint Conference on
  Natural Language Processing (EMNLP-IJCNLP)}, pages 540--551, Hong Kong,
  China. Association for Computational Linguistics.

\bibitem[{Laokulrat et~al.(2013)Laokulrat, Miwa, Tsuruoka, and
  Chikayama}]{laokulrat:SemEval:2013}
Natsuda Laokulrat, Makoto Miwa, Yoshimasa Tsuruoka, and Takashi Chikayama.
  2013.
\newblock \href {https://aclanthology.org/S13-2015} {{UTT}ime: Temporal
  relation classification using deep syntactic features}.
\newblock In \emph{Second Joint Conference on Lexical and Computational
  Semantics (*{SEM}), Volume 2: Proceedings of the Seventh International
  Workshop on Semantic Evaluation ({S}em{E}val 2013)}, pages 88--92, Atlanta,
  Georgia, USA. Association for Computational Linguistics.

\bibitem[{Laskar et~al.(2020)Laskar, Huang, and
  Hoque}]{laskar-etal-2020-contextualized}
Md~Tahmid~Rahman Laskar, Jimmy~Xiangji Huang, and Enamul Hoque. 2020.
\newblock \href {https://aclanthology.org/2020.lrec-1.676} {Contextualized
  embeddings based transformer encoder for sentence similarity modeling in
  answer selection task}.
\newblock In \emph{Proceedings of the Twelfth Language Resources and Evaluation
  Conference}, pages 5505--5514, Marseille, France. European Language Resources
  Association.

\bibitem[{Lee et~al.(2022)Lee, Lee, Jang, and Yu}]{lee-etal-2022-toward}
Seonghyeon Lee, Dongha Lee, Seongbo Jang, and Hwanjo Yu. 2022.
\newblock \href {https://doi.org/10.18653/v1/2022.acl-long.412} {Toward
  interpretable semantic textual similarity via optimal transport-based
  contrastive sentence learning}.
\newblock In \emph{Proceedings of the 60th Annual Meeting of the Association
  for Computational Linguistics (Volume 1: Long Papers)}, pages 5969--5979,
  Dublin, Ireland. Association for Computational Linguistics.

\bibitem[{Levin et~al.(1999)}]{levin1999objecthood}
Beth Levin et~al. 1999.
\newblock Objecthood: An event structure perspective.
\newblock \emph{Proceedings of CLS}, 35(1):223--247.

\bibitem[{Loshchilov and Hutter(2017)}]{loshchilov2019adamw}
Ilya Loshchilov and Frank Hutter. 2017.
\newblock \href {https://doi.org/10.48550/ARXIV.1711.05101} {Decoupled weight
  decay regularization}.

\bibitem[{Ma et~al.(2022)Ma, Hiraoka, and Okazaki}]{ma-etal-2022-joint}
Youmi Ma, Tatsuya Hiraoka, and Naoaki Okazaki. 2022.
\newblock \href {https://doi.org/10.18653/v1/2022.spnlp-1.2} {Joint entity and
  relation extraction based on table labeling using convolutional neural
  networks}.
\newblock In \emph{Proceedings of the Sixth Workshop on Structured Prediction
  for NLP}, pages 11--21, Dublin, Ireland. Association for Computational
  Linguistics.

\bibitem[{Manning et~al.(2014)Manning, Surdeanu, Bauer, Finkel, Bethard, and
  McClosky}]{manning:ACL:2014}
Christopher Manning, Mihai Surdeanu, John Bauer, Jenny Finkel, Steven Bethard,
  and David McClosky. 2014.
\newblock \href {https://doi.org/10.3115/v1/P14-5010} {The {S}tanford
  {C}ore{NLP} natural language processing toolkit}.
\newblock In \emph{Proceedings of 52nd Annual Meeting of the Association for
  Computational Linguistics: System Demonstrations}, pages 55--60, Baltimore,
  Maryland. Association for Computational Linguistics.

\bibitem[{Mariko et~al.(2022)Mariko, Abi-Akl, Trottier, and
  El-Haj}]{mariko-etal-2022-financial}
Dominique Mariko, Hanna Abi-Akl, Kim Trottier, and Mahmoud El-Haj. 2022.
\newblock \href {https://aclanthology.org/2022.fnp-1.16} {The financial
  causality extraction shared task ({F}in{C}ausal 2022)}.
\newblock In \emph{Proceedings of the 4th Financial Narrative Processing
  Workshop @LREC2022}, pages 105--107, Marseille, France. European Language
  Resources Association.

\bibitem[{Mariko et~al.(2020)Mariko, Labidurie, Ozturk, Akl, and
  de~Mazancourt}]{mariko:arXiv:2020}
Dominique Mariko, Estelle Labidurie, Yagmur Ozturk, Hanna~Abi Akl, and Hugues
  de~Mazancourt. 2020.
\newblock \href {https://doi.org/10.48550/ARXIV.2012.02498} {Data processing
  and annotation schemes for fincausal shared task}.

\bibitem[{Mirza and Tonelli(2016{\natexlab{a}})}]{mirza-tonelli:COLING:2016}
Paramita Mirza and Sara Tonelli. 2016{\natexlab{a}}.
\newblock \href {https://aclanthology.org/C16-1007} {{CATENA}: {CA}usal and
  {TE}mporal relation extraction from {NA}tural language texts}.
\newblock In \emph{Proceedings of {COLING} 2016, the 26th International
  Conference on Computational Linguistics: Technical Papers}, pages 64--75,
  Osaka, Japan. The COLING 2016 Organizing Committee.

\bibitem[{Mirza and Tonelli(2016{\natexlab{b}})}]{mirza-tonelli-2016-catena}
Paramita Mirza and Sara Tonelli. 2016{\natexlab{b}}.
\newblock \href {https://aclanthology.org/C16-1007} {{CATENA}: {CA}usal and
  {TE}mporal relation extraction from {NA}tural language texts}.
\newblock In \emph{Proceedings of {COLING} 2016, the 26th International
  Conference on Computational Linguistics: Technical Papers}, pages 64--75,
  Osaka, Japan. The COLING 2016 Organizing Committee.

\bibitem[{Miwa and Sasaki(2014)}]{miwa-sasaki-2014-modeling}
Makoto Miwa and Yutaka Sasaki. 2014.
\newblock \href {https://doi.org/10.3115/v1/D14-1200} {Modeling joint entity
  and relation extraction with table representation}.
\newblock In \emph{Proceedings of the 2014 Conference on Empirical Methods in
  Natural Language Processing ({EMNLP})}, pages 1858--1869, Doha, Qatar.
  Association for Computational Linguistics.

\bibitem[{Ning et~al.(2018)Ning, Wu, and Roth}]{Ning:ACL:2018}
Qiang Ning, Hao Wu, and Dan Roth. 2018.
\newblock \href {https://doi.org/10.18653/v1/P18-1122} {A multi-axis annotation
  scheme for event temporal relations}.
\newblock In \emph{Proceedings of the 56th Annual Meeting of the Association
  for Computational Linguistics (Volume 1: Long Papers)}, pages 1318--1328,
  Melbourne, Australia. Association for Computational Linguistics.

\bibitem[{Pan and Yang(2010)}]{pan-transfer}
Sinno~Jialin Pan and Qiang Yang. 2010.
\newblock \href {https://doi.org/10.1109/TKDE.2009.191} {A survey on transfer
  learning}.
\newblock \emph{IEEE Transactions on Knowledge and Data Engineering},
  22(10):1345--1359.

\bibitem[{Papanikolaou et~al.(2019)Papanikolaou, Roberts, and
  Pierleoni}]{papanikolaou-etal-2019-deep}
Yannis Papanikolaou, Ian Roberts, and Andrea Pierleoni. 2019.
\newblock \href {https://doi.org/10.18653/v1/D19-6108} {Deep bidirectional
  transformers for relation extraction without supervision}.
\newblock In \emph{Proceedings of the 2nd Workshop on Deep Learning Approaches
  for Low-Resource NLP (DeepLo 2019)}, pages 67--75, Hong Kong, China.
  Association for Computational Linguistics.

\bibitem[{Pradhan et~al.(2013)Pradhan, Moschitti, Xue, Ng, Bj{\"o}rkelund,
  Uryupina, Zhang, and Zhong}]{pradhan2013towards}
Sameer Pradhan, Alessandro Moschitti, Nianwen Xue, Hwee~Tou Ng, Anders
  Bj{\"o}rkelund, Olga Uryupina, Yuchen Zhang, and Zhi Zhong. 2013.
\newblock \href {https://aclanthology.org/W13-3516} {Towards robust linguistic
  analysis using {O}nto{N}otes}.
\newblock In \emph{Proceedings of the Seventeenth Conference on Computational
  Natural Language Learning}, pages 143--152, Sofia, Bulgaria. Association for
  Computational Linguistics.

\bibitem[{Pradhan et~al.(2012)Pradhan, Moschitti, Xue, Uryupina, and
  Zhang}]{pradhan-etal-2012-conll}
Sameer Pradhan, Alessandro Moschitti, Nianwen Xue, Olga Uryupina, and Yuchen
  Zhang. 2012.
\newblock \href {https://aclanthology.org/W12-4501} {{C}o{NLL}-2012 shared
  task: Modeling multilingual unrestricted coreference in {O}nto{N}otes}.
\newblock In \emph{Joint Conference on {EMNLP} and {C}o{NLL} - Shared Task},
  pages 1--40, Jeju Island, Korea. Association for Computational Linguistics.

\bibitem[{Pustejovsky et~al.(2003{\natexlab{a}})Pustejovsky, Hanks, Sauri, See,
  Gaizauskas, Setzer, Radev, Sundheim, Day, Ferro
  et~al.}]{pustejovsky2003timebank}
James Pustejovsky, Patrick Hanks, Roser Sauri, Andrew See, Robert Gaizauskas,
  Andrea Setzer, Dragomir Radev, Beth Sundheim, David Day, Lisa Ferro, et~al.
  2003{\natexlab{a}}.
\newblock The timebank corpus.
\newblock In \emph{Proceedings of Corpus linguistics}, volume 2003, page~40,
  Lancaster, UK. Corpus linguistics.

\bibitem[{Pustejovsky et~al.(2003{\natexlab{b}})Pustejovsky, Hanks, Saurí,
  See, Gaizauskas, Setzer, Radev, Sundheim, Day, Ferro, and
  Lazo}]{pustejovsky:book:2003}
James Pustejovsky, Patrick Hanks, Roser Saurí, Andrew See, Rob Gaizauskas,
  Andrea Setzer, Dragomir Radev, Beth Sundheim, David Day, Lisa Ferro, and
  Marcia Lazo. 2003{\natexlab{b}}.
\newblock The timebank corpus.
\newblock \emph{Proceedings of Corpus Linguistics}, 2003:40.

\bibitem[{Radford et~al.(2019)Radford, Wu, Child, Luan, Amodei, Sutskever
  et~al.}]{radford2019language}
Alec Radford, Jeffrey Wu, Rewon Child, David Luan, Dario Amodei, Ilya
  Sutskever, et~al. 2019.
\newblock Language models are unsupervised multitask learners.
\newblock \emph{OpenAI blog}, 1(8):9.

\bibitem[{Rappaport~Hovav et~al.(2010)Rappaport~Hovav, Doron, and
  Sichel}]{hovav2010lexical}
Malka Rappaport~Hovav, Edit Doron, and Ivy Sichel. 2010.
\newblock \href {https://doi.org/10.1093/acprof:oso/9780199544325.001.0001}
  {\emph{Lexical Semantics, Syntax, and Event Structure}}.
\newblock Oxford University Press, UK.

\bibitem[{Sanh et~al.(2019)Sanh, Debut, Chaumond, and
  Wolf}]{sanh2019distilbert}
Victor Sanh, Lysandre Debut, Julien Chaumond, and Thomas Wolf. 2019.
\newblock \href {https://doi.org/10.48550/ARXIV.1910.01108} {Distilbert, a
  distilled version of bert: smaller, faster, cheaper and lighter}.

\bibitem[{Seganti et~al.(2021)Seganti, Firlag, Skowronska, Satlawa, and
  Andruszkiewicz}]{Seganti2021MultilingualEA}
Alessandro Seganti, Klaudia Firlag, Helena Skowronska, Michal Satlawa, and
  Piotr Andruszkiewicz. 2021.
\newblock \href {https://www.aclweb.org/anthology/2021.eacl-main.166/}
  {Multilingual entity and relation extraction dataset and model}.
\newblock In \emph{Proceedings of the 16th Conference of the European Chapter
  of the Association for Computational Linguistics: Main Volume, EACL 2021,
  Online, April 19 - 23, 2021}, online. Association for Computational
  Linguistics.

\bibitem[{Tan et~al.(2022)Tan, Hürriyetoğlu, Caselli, Oostdijk, Nomoto,
  Hettiarachchi, Ameer, Uca, Liza, and Hu}]{tan:CNC:2022}
Fiona~Anting Tan, Ali Hürriyetoğlu, Tommaso Caselli, Nelleke Oostdijk,
  Tadashi Nomoto, Hansi Hettiarachchi, Iqra Ameer, Onur Uca, Farhana~Ferdousi
  Liza, and Tiancheng Hu. 2022.
\newblock \href {https://doi.org/10.48550/ARXIV.2204.11714} {The causal news
  corpus: Annotating causal relations in event sentences from news}.

\bibitem[{Vaswani et~al.(2017)Vaswani, Shazeer, Parmar, Uszkoreit, Jones,
  Gomez, Kaiser, and Polosukhin}]{vaswani2017attention}
Ashish Vaswani, Noam Shazeer, Niki Parmar, Jakob Uszkoreit, Llion Jones,
  Aidan~N. Gomez, Lukasz Kaiser, and Illia Polosukhin. 2017.
\newblock \href {https://doi.org/10.48550/ARXIV.1706.03762} {Attention is all
  you need}.

\bibitem[{Yang et~al.(2022)Yang, Wang, Wu, Yang, and Zhang}]{yang:arXiv:2022}
Linyi Yang, Zhen Wang, Yuxiang Wu, Jie Yang, and Yue Zhang. 2022.
\newblock \href {https://doi.org/10.48550/ARXIV.2204.07408} {Towards
  fine-grained causal reasoning and qa}.

\bibitem[{Yang et~al.(2019)Yang, Dai, Yang, Carbonell, Salakhutdinov, and
  Le}]{yang2019xlnet}
Zhilin Yang, Zihang Dai, Yiming Yang, Jaime Carbonell, Ruslan Salakhutdinov,
  and Quoc~V. Le. 2019.
\newblock \href {https://doi.org/10.48550/ARXIV.1906.08237} {Xlnet: Generalized
  autoregressive pretraining for language understanding}.

\bibitem[{Zhang et~al.(2022)Zhang, Xia, Zhou, Jiang, Fu, and
  Zhang}]{zhang-etal-2022-crfsrl}
Yu~Zhang, Qingrong Xia, Shilin Zhou, Yong Jiang, Guohong Fu, and Min Zhang.
  2022.
\newblock \href {https://aclanthology.org/2022.coling-1.370} {Semantic role
  labeling as dependency parsing: Exploring latent tree structures inside
  arguments}.
\newblock In \emph{Proceedings of the 29th International Conference on
  Computational Linguistics}, pages 4212--4227, Gyeongju, Republic of Korea.
  International Committee on Computational Linguistics.

\bibitem[{Zhu et~al.(2022)Zhu, Dai, Su, Ma, Liu, Cai, Xiao, and
  Zhang}]{bars-sigir-2022}
Jieming Zhu, Quanyu Dai, Liangcai Su, Rong Ma, Jinyang Liu, Guohao Cai,
  Xi~Xiao, and Rui Zhang. 2022.
\newblock \href {https://doi.org/10.48550/ARXIV.2205.09626} {Bars: Towards open
  benchmarking for recommender systems}.

\bibitem[{Zhuang et~al.(2021)Zhuang, Wayne, Ya, and Jun}]{liu2019roberta}
Liu Zhuang, Lin Wayne, Shi Ya, and Zhao Jun. 2021.
\newblock \href {https://aclanthology.org/2021.ccl-1.108} {A robustly optimized
  {BERT} pre-training approach with post-training}.
\newblock In \emph{Proceedings of the 20th Chinese National Conference on
  Computational Linguistics}, pages 1218--1227, Huhhot, China. Chinese
  Information Processing Society of China.

\end{thebibliography}
\bibliographystyle{acl_natbib}

\appendix



\end{document}